\documentclass[a4paper,fleqn]{cas-dc}

\usepackage[numbers]{natbib}

\usepackage{algorithm}
\usepackage{algorithmic}

\usepackage{subcaption}

\def\tsc#1{\csdef{#1}{\textsc{\lowercase{#1}}\xspace}}
\tsc{WGM}
\tsc{QE}
\begin{document}
\let\WriteBookmarks\relax
\def\floatpagepagefraction{1}
\def\textpagefraction{.001}

% Short title
\shorttitle{FedIA: Instance-Aware Cross-Modal Representations for Federated Prompt Learning}    

% Short author
\shortauthors{X. Zhang et al.}    

% Main title of the paper
\title [mode = title]{FedIA: Instance-Aware Cross-Modal Representations for Federated Prompt Learning}

\author[1]{Xuanjing Zhang}%[<options>]

% Address/affiliation
\affiliation[1]{organization={School of Computer Science and Technology, Xi’an Jiaotong University},
            %addressline={}, 
            city={Xi'an},
%          citysep={}, % Uncomment if no comma needed between city and postcode
            postcode={710049}, 
            state={Shaanxi},
            country={China}}

% Address/affiliation
\affiliation[2]{organization={Air Traffic Control and Navigation School, Air Force Engineering University},
            %addressline={}, 
            city={Xi'an},
%          citysep={}, % Uncomment if no comma needed between city and postcode
            postcode={710051}, 
            state={Shaanxi},
            country={China}}

\author[2]{Qi Guo}%[]

\author[1]{Yanan Qiao}%[]
\cormark[1]
\ead{qiaoyanan@xjtu.edu.cn}
\cortext[1]{Corresponding author}

\author[1]{Yong Han}

\author[1]{Junge Bo}

\begin{abstract}
Federated Prompt Learning (FPL) enables parameter-efficient adaptation of pre-trained Vision-Language Models across decentralized clients without exchanging raw data. To cope with cross-client heterogeneity, existing FPL methods typically pair a globally shared prompt with a client-specific local prompt. However, this design personalizes only at the client level: the same local prompt is applied to all samples of a client, and the global and local prompts are combined in an input-agnostic, fixed manner. Consequently, it overlooks the substantial visual variations among samples within the same client and cannot adjust the balance between shared and personalized knowledge to each input. To address these limitations, we propose Federated Prompt Learning with Instance-Aware Cross-Modal Representations (FedIA), an FPL framework that learns instance-aware cross-modal representations. Specifically, FedIA first uses global text prompts as client-invariant semantic anchors to select image patches most relevant to each candidate class, suppressing distracting regions. The selected visual evidence then conditions the generation of instance-aware local text representations. Finally, an adaptive global--local fusion module calibrates the contributions of local and global text representations for each image--class pair. In this way, FedIA shifts personalization in FPL from the client level to the instance level while preserving the semantic knowledge shared across clients. Extensive experiments under domain shift, pathological label skew, and Dirichlet-based label skew show that FedIA outperforms FPL baselines, with maximum accuracy gains of 1.44\%--4.37\% over the latest baseline across the three settings. Ablation studies and analyses further verify the complementary contributions of instance-aware representation generation and adaptive global--local fusion.
\end{abstract}

% Keywords
% Each keyword is seperated by \sep
\begin{keywords}
Federated Learning \sep Prompt Learning\sep Vision-Language Models\sep
\end{keywords}

\maketitle

\section{Introduction}
Federated Prompt Learning (FPL)~\cite{DBLP:journals/tmc/GuoGWTX24_promptfl, DBLP:conf/icassp/ZhaoDLLL23_FedPrompt} has emerged as an efficient paradigm for adapting large-scale pre-trained Vision-Language Models (VLMs), such as CLIP~\cite{DBLP:conf/icml/RadfordKHRGASAM21_CLIP} and ALIGN~\cite{DBLP:conf/icml/JiaYXCPPLSLD21_ALIGN}, to decentralized data.
Rather than fine-tuning the entire model, FPL keeps the pre-trained backbone frozen and lets clients collaboratively optimize only a small set of learnable prompt tokens~\cite{DBLP:journals/tmc/GuoGWTX24_promptfl, DBLP:journals/ijcv/ZhouYLL22_CoOp} under the Federated Learning (FL) framework~\citep{DBLP:conf/aistats/McMahanMRHA17}.
By exchanging only these lightweight prompts, FPL leverages the transferable knowledge of VLMs while keeping private data local and substantially reducing computation and communication costs.

Despite these advantages, FPL faces a central challenge arising from data heterogeneity, as client data are typically non-independent and identically distributed (Non-IID)~\cite{DBLP:journals/ijon/GuoQQWL23_fedmcsa, DBLP:conf/cvpr/HuangYSL023_FPL, DBLP:conf/icml/KarimireddyKMRS20_scaffold, DBLP:conf/mlsys/LiSZSTS20_fedprox}. 
Different clients possess data from distinct domains, classes, or statistical distributions, making it difficult for a shared prompt to adequately capture both common semantic knowledge and client-specific characteristics.

\begin{figure}[pos=t]
    \centering
    \includegraphics[width=\columnwidth]{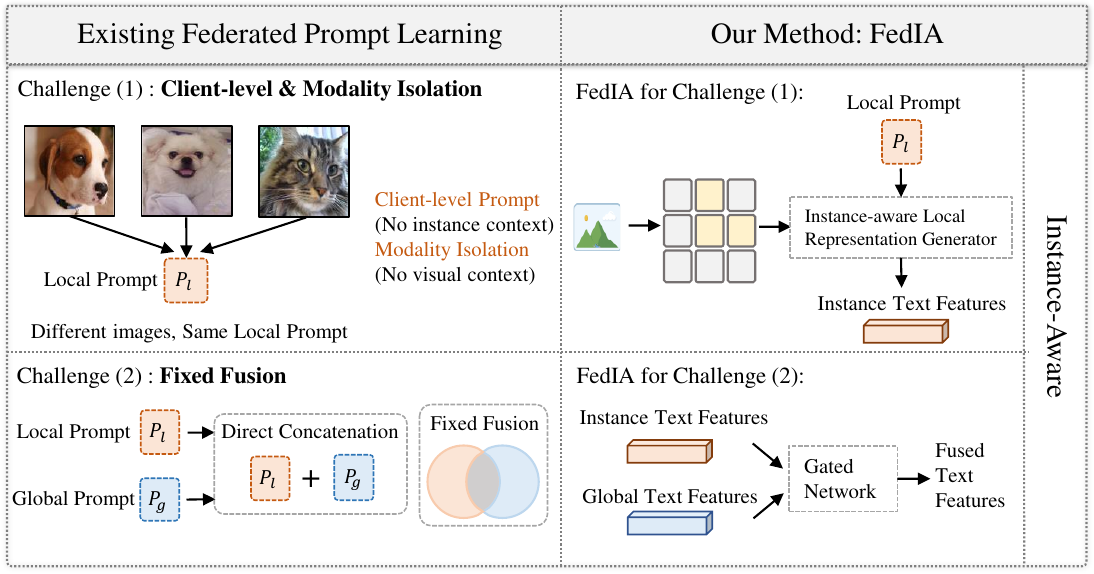}
    \caption{Motivation and overview of FedIA. Challenge (1): Existing FPL methods typically use client-level prompts, which lack visual context and fail to capture instance-level variations within each client. Challenge (2): They also rely on input-agnostic fixed global--local fusion, making it difficult to adaptively balance personalization and generalization.}
    \label{intro}
\end{figure}

To address client heterogeneity, existing FPL methods commonly maintain a globally shared prompt to capture knowledge common across clients and a personalized local prompt to model client-specific characteristics~\cite{DBLP:conf/icml/CuiLW024_FedPGP, DBLP:conf/icml/Fang0WYY25_FedPHA, DBLP:conf/www/Guo0W23_pFedPrompt}. 
The global prompt promotes knowledge transfer and generalization across heterogeneous clients, whereas the local prompt enables adaptation to distinctive data distributions and visual patterns. 
Although this global--local design improves personalization, it remains limited in two important aspects, as illustrated in Figure~\ref{intro}.

Existing FPL methods often rely on client-level local prompts, applying the same prompt parameters to all samples within a client~\cite{DBLP:conf/icml/Fang0WYY25_FedPHA, DBLP:conf/cvpr/LiHW024_FedOTP, DBLP:conf/mm/LiZZY025_FedGR, DBLP:conf/iclr/Luo0W25_pFedMoAP}.
While this design can capture coarse client-level biases, it cannot adapt to the instance-level visual variations that naturally arise among samples from the same client.
For example, even samples from one client differ in background context, visual style, or class-relevant regions. 
Client-level prompts ignore these differences because they are optimized independently from the visual modality and do not dynamically adjust their semantic focus according to each input image. 
This modality isolation~\cite{DBLP:conf/iccv/ZhengYHYH25_HiCroPL} weakens fine-grained cross-modal alignment and causes the local prompt to encode only averaged client preferences rather than discriminative instance-specific cues.
Moreover, making local representations sensitive to individual images introduces a related coordination challenge: the contribution of local visual evidence may vary across samples, requiring the interaction with shared semantic knowledge to adapt accordingly.
However, many existing global--local designs often combine global and local prompts in an input-agnostic manner, such as direct concatenation or fixed fusion~\cite{DBLP:conf/icml/CuiLW024_FedPGP, DBLP:conf/aaai/ChenLWCZFXS26_FaST-PT, DBLP:conf/icml/YangHW0Y25_FedDDA}. 
These strategies assume that the relative contributions of global and local knowledge remain constant across all samples, which fails to account for the varying semantic requirements of different images. 
Some images may benefit more from global knowledge that captures class semantics shared across clients, whereas others may require stronger local adaptation to represent domain-specific appearances or client-level characteristics. Consequently, a fixed fusion strategy can assign excessive weight to local prompts, weakening the consistency of shared semantic information, or rely too heavily on global prompts, limiting the expression of personalized visual cues. 
These observations suggest that effective FPL methods should generate fine-grained text representations enriched with visual information and dynamically coordinate them with globally shared semantic knowledge~\cite{DBLP:conf/icml/CuiLW024_FedPGP, DBLP:conf/cvpr/DengTL24_SGPT}.

In this work, we propose \textbf{Fed}erated Prompt Learning with \textbf{I}nstance-\textbf{A}ware Cross-Modal Representations, termed \textbf{FedIA}, which learns cross-modal local prompt representations and adaptively coordinates them with globally shared semantic knowledge for each image--class pair. 
FedIA is designed to address the limited instance-level adaptability of existing FPL methods and the fixed coordination between global and local representations under heterogeneous client distributions.
To address the first challenge, FedIA introduces Instance-aware Local Representation Generation (ILRG), which dynamically generates local text representations conditioned on image semantics instead of assigning a fixed prompt to all samples from the same client. 
To strengthen fine-grained cross-modal alignment, FedIA further employs Global Prompt-Guided Patch Selection, which uses global text prompts as semantic anchors to prioritize class-relevant visual patches while reducing the influence of patches with low relevance to the candidate classes.
The selected visual evidence is then injected into the instance-aware local text representation generation process, enabling FedIA to construct instance-aware cross-modal representations that capture discriminative visual semantics while remaining aligned with textual class knowledge.
To address the second challenge, FedIA designs Adaptive Global--Local Representation Fusion (AGLRF), which adaptively calibrates the contributions of instance-aware local text representations and global text representations according to the current image and class semantics. Through these components, FedIA jointly improves instance-level personalization and global--local text representation coordination in heterogeneous federated prompt learning.

Our main contributions are summarized as follows:
\begin{itemize}
  \item We expose an underexplored instance-level bottleneck in personalized federated prompt learning: a shared local prompt is used to represent all samples within each client, while global and local knowledge are integrated in an input-agnostic manner. We build on this observation to shift the focus of personalization from modeling coarse client-level distributions to capturing fine-grained, sample-dependent visual semantics.
  \item We propose FedIA, a unified FPL framework that generates instance-aware local text representations guided by selected class-relevant visual patches and adaptively calibrates personalization with globally shared knowledge for each input.
  \item Extensive experiments on three federated settings covering domain and label-distribution heterogeneity show that FedIA generally outperforms representative FPL methods. Further analyses confirm the complementary benefits of instance-aware text representation and adaptive global--local fusion.
\end{itemize}

%%%%%%%%%%%%%%%%%%%%%%%%%%%%%%%%%%%%%%%%%%%%%%%%%%%%%%%%%%%%%%%%%%%%%%%

\section{Related Work}
\subsection{Federated Learning}
Federated Learning (FL) has emerged as a promising paradigm to enable collaborative model training across multiple participants without sharing their raw data, thereby effectively preserving data privacy. 
A foundational algorithm FedAvg~\cite{DBLP:conf/aistats/McMahanMRHA17} aggregates the locally trained model parameters from all participants to update a shared global model. 
While FedAvg is simple and effective under independent and identically distributed (IID) data settings, its performance deteriorates significantly when faced with Non-IID data distributions where participants often hold data from different domains, classes, or statistical distributions~\cite{DBLP:conf/mlsys/LiSZSTS20_fedprox}. 
To address this challenge, numerous methods have been proposed to mitigate the adverse effects of heterogeneity~\cite{DBLP:conf/icml/KarimireddyKMRS20_scaffold, DBLP:conf/mlsys/LiSZSTS20_fedprox}.
FedProx~\cite{DBLP:conf/mlsys/LiSZSTS20_fedprox} makes local updates close to the global model by adding a proximal term to the local objective.
Although these methods have made some progress, they still distribute a unified global model which is difficult to apply to all participants.

To address the limitations of a single global model under heterogeneous client data distributions, personalized federated learning (PFL)~\cite{DBLP:journals/ijon/GuoQQWL23_fedmcsa, DBLP:journals/tnn/TanYCY23_towards} has been proposed. 
PFL aims to learn a personalized model tailored to each participant's local data distribution while benefiting from knowledge shared across participants. 
Existing approaches achieve this objective through personalized optimization, model partitioning, or adaptive aggregation. 
FedPer~\cite{DBLP:journals/corr/abs-1912-00818_fedper} separates the model into shared base layers trained through federated averaging and personalized layers retained locally. 

Despite their effectiveness in addressing data heterogeneity, conventional FL and PFL methods often require substantial model training and parameter exchange. 
Leveraging pre-trained models offers an alternative by transferring existing knowledge to downstream tasks, reducing the need to learn representations from scratch. 
However, updating and communicating all parameters of large pre-trained models remains costly in federated settings. 
This motivates federated parameter-efficient fine-tuning, which enables privacy-preserving adaptation of pre-trained models to downstream tasks without sharing raw data, while updating and exchanging only a small set of trainable parameters and keeping the backbone frozen.

\subsection{Prompt Learning in VLMs}
Large-scale vision-language models (VLMs)~\cite{DBLP:conf/icml/RadfordKHRGASAM21_CLIP, DBLP:conf/icml/JiaYXCPPLSLD21_ALIGN, DBLP:conf/iclr/YaoHHLNXLLJX22_filip}, such as CLIP~\citep{DBLP:conf/icml/RadfordKHRGASAM21_CLIP}, learn transferable cross-modal representations from massive collections of image-text pairs and exhibit strong zero-shot generalization.
By aligning visual and textual representations in a shared embedding space, these models enable recognition through similarity matching between image embeddings and textual class descriptions.
This alignment provides a foundation for adapting pre-trained VLMs to a wide range of downstream recognition tasks.
To achieve efficient adaptation, prompt learning has emerged as an effective parameter-efficient fine-tuning paradigm~\cite{DBLP:conf/aaai/ChenZJHW26_dropout, DBLP:conf/eccv/JiaTCCBHL22_VPT, DBLP:conf/iccv/KhattakWNK0K23_PromptSRC, DBLP:conf/aaai/LiMCDLL26_GuiDG, DBLP:journals/ijcv/SongLLWZXX26_AmPLe, DBLP:conf/icml/WangCFZDW24_SPT, DBLP:conf/cvpr/YaoZX23_KgCoOp, DBLP:journals/kbs/ZhangLY26_hckp}.
It optimizes a small set of prompt parameters while keeping the pre-trained backbone frozen, enabling task-specific adaptation with a limited number of trainable parameters.
CoOp~\cite{DBLP:journals/ijcv/ZhouYLL22_CoOp} replaces hand-crafted text templates with learnable context vectors, allowing task-specific contextual information to be learned directly from downstream data.
To better preserve the generalization ability of the pre-trained model, KgCoOp~\cite{DBLP:conf/cvpr/YaoZX23_KgCoOp} minimizes the discrepancy between text embeddings generated by learned and hand-crafted prompts.
This constraint preserves pre-trained textual knowledge and improves generalization to unseen classes.
Beyond text-side adaptation, MaPLe~\cite{DBLP:conf/cvpr/KhattakR0KK23_MaPLe} extends prompt learning to both the textual and visual branches through cross-modal coupling, promoting coordinated adaptation across modalities at multiple layers.

These methods demonstrate the effectiveness of prompt learning for adapting pre-trained VLMs while preserving their transferable knowledge.
However, they are primarily designed for centralized training and do not explicitly account for data distributed across heterogeneous clients.
Extending prompt learning to such settings requires accommodating differences in client data while enabling collaborative knowledge sharing.
This motivates federated prompt learning, which combines efficient VLMs adaptation with collaborative learning under decentralized data distributions.

\subsection{Federated Prompt Learning}
To adapt vision-language models (VLMs) to decentralized scenarios, federated prompt learning (FPL) has attracted increasing attention~\cite{DBLP:conf/cvpr/FengLX0FZ23_FedPR, DBLP:conf/aaai/SuY0X24_FedAPT, DBLP:conf/ijcai/WangHCSWQYWLY25_empirical, DBLP:conf/infocom/ZengWSL26_FT_PromptFL, DBLP:journals/asc/ZhangZXMWZY26_FedHAP, DBLP:conf/cvpr/ZhuBGRYL26_COTE}.
By optimizing a small set of prompt parameters while keeping the pre-trained VLMs backbone frozen, FPL substantially reduces the number of trainable parameters and the communication overhead associated with federated adaptation.
However, heterogeneity in local domains, class distributions, and visual characteristics poses challenges to collaborative learning.
An important objective is therefore to preserve transferable semantic knowledge across clients while accommodating their specific adaptation needs.

PromptFL~\cite{DBLP:journals/tmc/GuoGWTX24_promptfl} introduces prompt tuning into federated learning, allowing clients to upload locally optimized prompts for server-side aggregation.
Through this process, clients collaboratively acquire transferable task knowledge without sharing raw data.
Building on this paradigm, subsequent methods explore conditional prompt generation and personalized prompt learning to improve generalization and local adaptation.
FedTPG~\cite{DBLP:conf/iclr/QiuLMGLPL24_FedTPG} collaboratively learns a shared prompt generator conditioned on task-related textual information.
This design enables task-aware prompt generation and improves generalization to unseen classes and datasets.
To explicitly balance shared knowledge with client-specific adaptation, several methods incorporate personalized components into federated prompt learning.
FedPGP~\cite{DBLP:conf/icml/CuiLW024_FedPGP} personalizes a globally shared prompt through client-specific low-rank adaptation and introduces a prompt-wise contrastive loss to balance generalization and personalization.
FedOTP~\cite{DBLP:conf/cvpr/LiHW024_FedOTP} combines global and local prompts to capture shared knowledge and client-specific category characteristics, respectively.
It further employs unbalanced optimal transport to align their textual representations with local visual features.
FedPHA~\cite{DBLP:conf/icml/Fang0WYY25_FedPHA} combines fixed-length global prompts with variable-length local prompts and uses SVD-based projection to mitigate conflicts between global and local knowledge.
Together, these methods highlight the importance of coordinating shared and personalized knowledge under client heterogeneity.

Unlike these methods, FedIA does not treat the local prompt as a client-level representation.
Instead, it emphasizes instance-level adaptation by generating local text representations conditioned on selected visual patches from each image.
These image-conditioned representations are then adaptively fused with global text representations for each image--class pair.
This design allows the contribution of local visual evidence to vary across both images and candidate classes, integrating instance-specific visual information with shared semantic knowledge.

\section{Preliminaries}
\subsection{Prompt Learning}
Large-scale pre-trained vision-language models, such as CLIP \cite{DBLP:conf/icml/RadfordKHRGASAM21_CLIP}, demonstrate strong cross-modal representation and zero-shot transferability through contrastive learning on massive image-text pairs.
The CLIP architecture comprises an image encoder $f(\cdot)$ and a text encoder $g(\cdot)$. 
In conventional zero-shot classification, the model relies on manually crafted discrete prompt templates (e.g., "a photo of a [CLASS]"). 
For the $h$-th class $c_h$, the class name is embedded into the template and processed by the text encoder to generate the text embedding. Given an input image $x$, CLIP computes the cosine similarity between the visual features $f(x)$ and the embeddings of all candidate classes, ultimately selecting the class with the highest similarity score as the predicted result.

To enhance the adaptability of CLIP to downstream tasks, Parameter-Efficient Fine-Tuning methods optimize only a small subset of parameters while keeping the pre-trained backbone frozen. 
Among these methods, prompt learning is particularly suitable for federated adaptation due to its compact trainable parameterization and low communication overhead. 
Specifically, prompt learning replaces hand-crafted templates with learnable continuous context vectors, denoted as $P=\{p_1,p_2,...,p_L\} \in \mathbb{R}^{L \times d}$, where $L$ is the prompt length and $d$ is the embedding dimension.
The input to the text encoder for the $h$-th class, denoted by $P_h=\{p_1,...,p_L,c_h\}$, is formed by concatenating learnable context vectors with the class name embedding $c_h$.
Given an image sample $x$ with label $h$, the prediction probabilities are as follows:
\begin{equation}
\label{sim}
P(h|x)=\frac{exp(sim(f(x),g(P_h))/\tau)}{\sum_{c=1}^{N_c} exp(sim(f(x),g(P_c))/\tau)},
\end{equation}
where $N_c$ is the number of classes, $sim(\cdot,\cdot)$ is cosine similarity and $\tau$ is the temperature parameter.
The learnable prompts are optimized using the cross-entropy loss:
\begin{equation}
\label{ce_pl}
\mathcal{L}_{CE}=-\sum_{h=1}^{N_c}{y_h\ \log\ P(h|x)},
\end{equation}
where $y_h$ is the $h$-th component of the one-hot label vector.

\subsection{Federated Prompt Learning}
We consider a federated learning system consisting of $N$ clients and a central server. 
Each client $i$ is equipped with a pre-trained CLIP model and a local dataset $\mathcal{D}_i=\{(x_j,y_j)\}_{j=1}^{|\mathcal{D}_i|}$, where $(x_j,y_j)$ denotes the $j$-th image--label pair and $|\mathcal{D}_i|$ is the number of local samples. 
To accommodate client heterogeneity, our framework adopts a dual-prompt mechanism comprising a personalized local prompt that captures client-specific characteristics and a shared global prompt that captures common semantic knowledge across clients.
Let $\mathcal{C}_t$ be the selected clients at communication round $t$. 
For each $t$, the selected clients initialize their global prompts with the global prompt from the previous round. Subsequently, each client $i$ performs local training for $E$ epochs to optimize the local prompt and the global prompt by minimizing the cross-entropy loss. For each epoch $e$, the updates for the prompts are formulated as:
\begin{equation}
\label{PG}
P_{g,i}^{t,e+1}=P_{g,i}^{t,e}-\eta\nabla_{P_{g,i}}\mathcal{L}(P_{g,i}^{t,e},P_{l,i}^{t,e};\mathcal{D}_i),
\end{equation}
\begin{equation}
\label{PL}
P_{l,i}^{t,e+1}=P_{l,i}^{t,e}-\eta\nabla_{P_{l,i}}\mathcal{L}(P_{g,i}^{t,e},P_{l,i}^{t,e};\mathcal{D}_i),
\end{equation}
where $P_{g,i}$ denotes the global prompt and $P_{l,i}$ denotes the local prompt of client $i$.

After local training, the global prompt $P_{g,i}^{t,E}$ is uploaded to the server, while the local prompt $P_{l,i}^{t,E}$ is retained on the client to preserve personalized knowledge. The server then aggregates the global prompts received from all participating clients as follows:
\begin{equation}
\label{PG_agg}
P_{g}^{t} = \sum_{i \in \mathcal{C}_t}\frac{|\mathcal{D}_i|}{\sum_{j \in \mathcal{C}_t}|\mathcal{D}_j|}P_{g,i}^{t,E}.
\end{equation}
The server distributes the aggregated global prompt to the clients for the next communication round.

\begin{figure*}[pos=t]
    \centering
    \includegraphics[width=\textwidth]{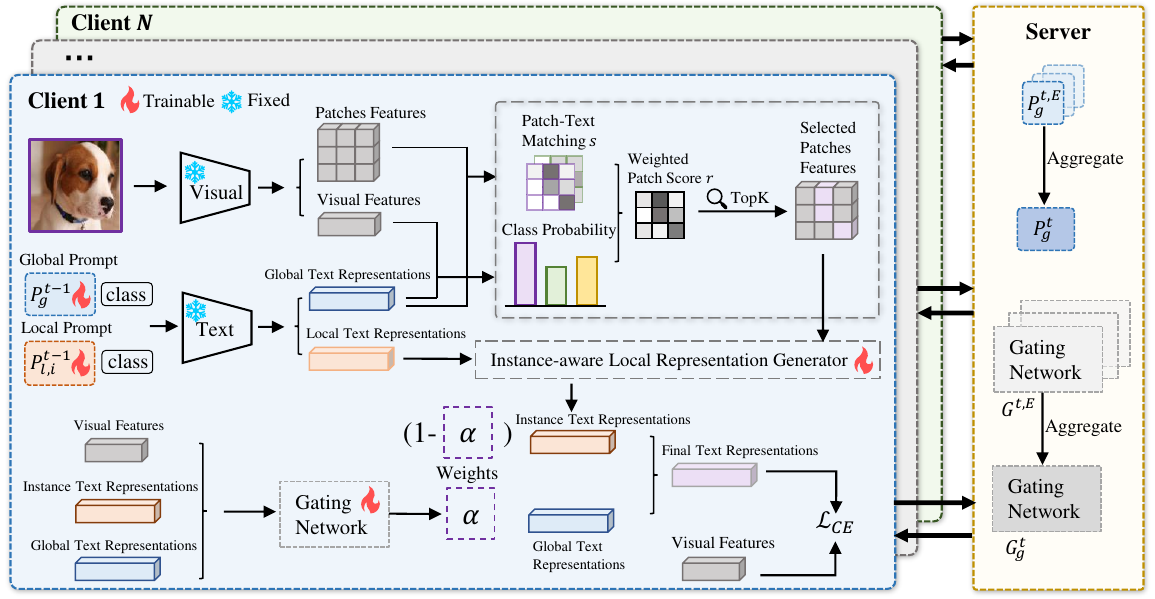}
    \caption{Framework of FedIA. FedIA first performs global-prompt-guided patch selection to retain class-relevant visual patches for each input image. The selected patches are then used by an instance-aware text representation generator to construct image-conditioned local representations. A lightweight gating network further adaptively fuses the instance text features with global knowledge, producing calibrated text representations balance personalized information and shared semantics.}
    \label{FedIA}
\end{figure*}

\section{Methodology}
In this section, we propose Federated Prompt Learning with Instance-Aware Cross-Modal Representations (FedIA), a novel FPL framework.
We leverage Instance-aware Local Text Representation Generation to capture personalized knowledge and Adaptive Global--Local Representation Fusion to calibrate it with global consensus.
The overview of FedIA is shown in Figure~\ref{FedIA}. 

\subsection{Instance-Aware Local Representation Generation}
In FL, data heterogeneity across clients remains a critical challenge. 
Existing FPL methods lack the fine-grained adaptability needed to capture instance-level variations and underexploit visual modality information.
To address this, we propose Instance-aware Local Representation Generation (ILRG), which generates instance-aware local text representations from visual semantics.
Because image-level features encompass visual cues with varying degrees of relevance to the target category, we further introduce Global Prompt-Guided Patch Selection to identify and retain patches that provide the most category-relevant information.

During local optimization for client $i$, for a given $j$-th input image $x_j$, let $z_j=\{z_{j,p}\}_{p=1}^{m} \in \mathbb{R}^{m \times d}$ denote the extracted visual patch features, obtained following \cite{DBLP:conf/cvpr/DongBZZCY00YC0Y23_Maskclip}, where $m$ is the number of patches and $d$ is the feature dimension. 
And $v_j=f(x_j) \in \mathbb{R}^{d}$ denotes the holistic visual features of the entire image. 
The global text representations are computed as $t_{g,i} = g(P_{g,i})$. 
We apply $L_2$ normalization to both the patch features and the global text representations to ensure scale invariance. 
Subsequently, the semantic relevance between each patch and the global text representation of each category is mathematically formulated as:
\begin{equation}
\label{patch_glo_cos}
s_{j,p,h}=cos(z_{j,p},t_{g,i,h})=\frac{z_{j,p}^{\top} t_{g,i,h}}{||z_{j,p}||_{2}||t_{g,i,h}||_2},
\end{equation}
where $s_{j,p,h}$ quantifies the semantic similarity between the $p$-th patch feature of the $j$-th image and the global text embeddings corresponding to the $h$-th class.
CLIP is pretrained with a contrastive objective that aligns semantically corresponding visual and textual representations in a shared embedding space. Therefore, after visual projection and $L_2$ normalization, the cosine similarity between a patch feature and a class-conditioned global text representation can be used as a semantic compatibility score. Specifically, $s_{j,p,h}$ measures the degree to which the $p$-th patch of image $x_j$ is aligned with the semantic concept of class $h$.
The global prompt provides a client-invariant class prototype, while the patch features provide instance-level visual evidence.
Thus, global-prompt-guided patch selection enables the model to identify visual regions that are more semantically relevant to the target category.

Then, the image-level visual features and the global text representations are utilized to compute the global logits, which are passed through a softmax function to derive the class weights. 
These weights explicitly reflect the semantic distribution of the input image across the global prompt branch.
The mathematical formulation is presented as follows:
\begin{equation}
\label{glo_logits}
\sigma_{j,h} = \gamma sim(v_{j},t_{g,i,h}),
\end{equation}
\begin{equation}
\label{glo_logits_softmax}
\omega_{j,h} = \frac{exp(\sigma_{j,h})}{\sum_{c=1}^{N_c}exp(\sigma_{j,c})},
\end{equation}
where $\gamma$ is the logit scale of CLIP, $sim(\cdot,\cdot)$ represents cosine similarity, and $\omega_{j,h}$ denotes the prediction weight of the $j$-th image for the $h$-th class.

Following this, the patch-class similarity scores are weighted and aggregated using these weights, yielding the final importance score for each individual patch, which is mathematically defined as:
\begin{equation}
\label{r}
r_{j,p}=\sum_{h=1}^{N_c}\omega_{j,h}s_{j,p,h},
\end{equation}
where $r_{j,p}$ represents the final importance score of the $p$-th patch feature within the $j$-th image, collectively forming the image-specific patch importance $r_{j}\in \mathbb{R}^m$.
Consequently, the top-$k$ visual patches with the highest scores are retained:
\begin{equation}
\label{topk}
\mathcal{K}_j=TopK(r_{j},k),
\end{equation}
where $k$ is the number of patches to be retained and the selected set of patches is $z_j^*=\{z_{j,p}|p\in \mathcal{K}_j\} \in \mathbb{R}^{k \times d}$.

After selecting semantically relevant patches, we construct instance-specific local text representations that incorporate fine-grained visual information from each input image to better adapt to local data distributions.
For each client $i$, we introduce an Instance-aware Local Text Representation Generator, implemented as a cross-attention \cite{DBLP:conf/nips/VaswaniSPUJGKP17_attention} module with parameters $I_i$, to dynamically generate instance-aware representations by incorporating visual information from the selected class-relevant image patches. 
To generate instance-aware text representations, the Instance-aware Local Text Representation Generator uses text features to query relevant visual information from the selected image patches.
Specifically, we use the local text features $t_{l,i}=g(P_{l,i})$ as queries (Q) and the selected visual patch features $z_j^*$ as keys (K) and values (V).
Formally, the operation is defined as:
\begin{equation}
\label{cross-att}
t_{j}=\phi(t_{l,i},z_j^*)=MLP\left(softmax\left(\frac{QK^\top}{\sqrt{d_k}}\right)V\right),
\end{equation}
\begin{equation}
\label{qkv}
Q = t_{l,i} W_Q,\ K = z_j^* W_K,\ V = z_j^* W_V,
\end{equation}
where $W_Q \in \mathbb{R}^{d \times d_k}$, $W_K \in \mathbb{R}^{d \times d_k}$, $W_V \in \mathbb{R}^{d \times d_v}$ are learnable linear projection matrices, $d_k$ is the scaling factor representing the dimension of the keys.
The Instance-aware Local Text Representation Generator is updated locally on each client, learning to dynamically align visual semantics with text representations under client-specific heterogeneous data distributions.

%%%%%%%%%%%%%%%%%%%%%%%%%%% 

\begin{algorithm}[t]
\caption{Federated Prompt Learning with Instance-Aware Cross-Modal Representations (FedIA)}
\label{alg:FedIA}
\begin{algorithmic}[1]

\REQUIRE
Communication rounds $R$; Local epochs $E$; Learning rate $\eta$; Number of selected patches $k$; Client datasets $\{\mathcal{D}_i\}_{i=1}^{N}$; Selected client sets $\{\mathcal{C}_t\}_{t=1}^{R}$; Initial server parameters $P_g^0,G_g^0$; Initial local parameters $\{P_{l,i}^0,I_i^0\}_{i=1}^{N}$.

\ENSURE
Final server parameters $(P_g^R,G_g^R)$ and local parameters $\{P_{l,i}^R, I_i^R\}_{i=1}^{N}$.

\STATE \textit{// Federated Rounds}
\FOR{$t=1,2,\ldots,R$}

    \STATE \textit{// Local Training}
    \FOR{each client $i\in\mathcal{C}_t$ in parallel}
        \STATE $(P_{g,i}^{t},G_{g,i}^{t},P_{l,i}^{t},I_i^{t})$
        \STATE \hspace{1em}$\leftarrow
        \textbf{LocalUpdate}(P_g^{t-1},G_g^{t-1},
        P_{l,i}^{t-1},I_i^{t-1},\mathcal{D}_i)$
    \ENDFOR

    \STATE \textit{// Server Aggregation}

    \STATE $P_g^t\leftarrow
    \sum_{i\in\mathcal{C}_t}\dfrac{|\mathcal{D}_i|}
    {\sum_{j\in\mathcal{C}_t}|\mathcal{D}_j|}P_{g,i}^{t}$

    \STATE $G_g^t\leftarrow
    \sum_{i\in\mathcal{C}_t}\dfrac{|\mathcal{D}_i|}
    {\sum_{j\in\mathcal{C}_t}|\mathcal{D}_j|}G_{g,i}^{t}$

\ENDFOR

\STATE \textbf{return} $(P_g^R,G_g^R)$ and $\{P_{l,i}^R, I_{i}^R\}_{i=1}^{N}$.

\STATE
\STATE \textbf{LocalUpdate}$(P_g,G_g,P_{l,i},I_i,\mathcal{D}_i)$:

\STATE $P_{g,i}\leftarrow P_g,\quad G_{g,i}\leftarrow G_g$

\FOR{$e=1,2,\ldots,E$}
    \FOR{each mini-batch $B\subset\mathcal{D}_i$}

        \STATE Extract patch features $z_{j}$ and image features
        $v_j$ using the frozen image encoder; obtain
        $t_{g,i,h}$ and $t_{l,i,h}$ using the text encoder.

        \STATE Compute $s_{j,p,h}$, $\omega_{j,h}$, $r_{j,p}$,
        $\mathcal{K}_j$, and $z_j^*$ by Eqs.~\eqref{patch_glo_cos}--\eqref{topk}.

        \STATE Compute instance-aware text representations
        $t_{j,h}$ by Eqs.~\eqref{cross-att}--\eqref{qkv}.

        \STATE Compute $\alpha_{j,h}$, $\bar{t}_{j,h}$,
        class probabilities, and $\mathcal{L}_{CE}$ by Eqs.~\eqref{gate_input}--\eqref{ce_ins}.

        \STATE $\theta_i\leftarrow
        \theta_i-\eta\nabla_{\theta_i}\mathcal{L}_{CE}$,
        \quad
        $\theta_i=\{P_{g,i},P_{l,i},I_i,G_{g,i}\}$

    \ENDFOR
\ENDFOR

\STATE \textbf{return}
$(P_{g,i},G_{g,i},P_{l,i},I_i)$.

\end{algorithmic}
\end{algorithm}

%%%%%%%%%%%%%%%%%%%%%%%%%%%

\subsection{Adaptive Global--Local Representation Fusion}
Instance-aware local text representations capture class-relevant visual cues and adapt to individual samples, emphasizing client-specific patterns learned from local data. Global text representations derived from the shared prompt provide complementary class semantics shared across clients, while offering less information about image-specific visual details.
The relative importance of these two sources of information varies across images and categories. To coordinate their complementary strengths, we introduce Adaptive Global--Local Representation Fusion (AGLRF), which dynamically adjusts their contributions for each image--class pair.

Specifically, for the $j$-th image, $t_{j}$ denotes the instance-level text features. 
We concatenate the instance-aware local text representation, the global text representation and the image feature, and feed them into a lightweight gating network:
\begin{equation}
\label{gate_input}
\alpha_{j,h} = \sigma(MLP([t_{j,h};t_{g,i,h};v_j])),
\end{equation}
where $\alpha_{j,h}$ denotes the gating weight for input image $j$ and class $h$, 
$t_{j,h}$ is the instance-aware text representation for class $h$ generated by image $j$, 
$t_{g,i,h}$ is the global text representation of class $h$, 
$\sigma(\cdot)$ denotes the sigmoid activation function, 
and the gating network is an MLP consisting of two linear layers, with parameters represented by $G_i$.
The final text features are obtained by adaptively interpolating between the instance-aware local representation and the global representation:
\begin{equation}
\label{gate}
\bar{t}_{j,h}=(1-\alpha_{j,h})t_{j,h}+\alpha_{j,h}t_{g,i,h}.
\end{equation}

Rather than using a fixed fusion ratio, the gate takes the instance-aware text representation, the global text representation, and the image representation as input, and estimates which text representation is more reliable for each image--class pair.
The instance-aware text representation captures client- and image-specific characteristics, providing personalized evidence from the local data distribution. 
In contrast, the global representation serves as a stable semantic anchor shared across clients, capturing common semantic knowledge.
The image representation further provides complementary context about the input.
By jointly considering these three sources, the gate determines the relative contributions of local and global knowledge for each sample. 
This enables the model to adaptively emphasize distinctive instance-level evidence or rely on global semantics as needed, achieving a better balance between instance-level adaptation and global semantic consistency.

\subsection{Local Training and Server Aggregation} 
During local training, each client optimizes the prompt learner, instance-aware local text representation generator, and lightweight fusion module while keeping the CLIP image and text encoders frozen.
Given an input image, the model obtains the image features and patch-level visual tokens from the frozen CLIP visual encoder. 
The selected local patches are used to generate instance-aware text representations, which are then adaptively fused with the global text representations to obtain the final text representations $\bar{t}$. 
The loss function is formulated as:
\begin{equation}
\label{ce_ins}
\begin{aligned}
\mathcal{L}_{CE}=\frac{1}{B}\sum_{b=1}^B{\left(-\sum_{h}y_h\ \text{log}\ P_b(h|x_b)\right)}, \\
\text{where}\ P_b(h|x_b)=\frac{exp(sim(f(x_b),\bar{t}_{b,h})/\tau)}{\sum_{c=1}^{N_c} exp(sim(f(x_b),\bar{t}_{b,c})/\tau)}.
\end{aligned}
\end{equation}
After local training, the server aggregates global parameters, including global prompt $P_g$ and fusion network $G$:
\begin{equation}
\label{agg1}
P_g^{t} = \sum_{i \in \mathcal{C}_t} \frac{|\mathcal{D}_i|}{\sum_{j \in \mathcal{C}_t}|\mathcal{D}_j|} P_{g,i}^{t,E},
\end{equation}
\begin{equation}
\label{agg2}
G_g^{t}=\sum_{i \in \mathcal{C}_t} \frac{|\mathcal{D}_i|}{\sum_{j \in \mathcal{C}_t}|\mathcal{D}_j|} G_i^{t,E}.
\end{equation}
To distill global consensus, the central server aggregates the localized knowledge uploaded from all participating clients, thereby extracting a unified semantic representation. 
Upon completing this aggregation, the server broadcasts the global prompt and the updated gating network weights back to each client for the next round of training.

\section{Experiments}
\subsection{Experimental Setup}

\subsubsection{Datasets.}
We evaluate our method FedIA on multiple public benchmark datasets. We design three distinct Non-IID settings: 
\begin{itemize}
    \item Domain Shift: We utilize three multi-domain datasets, PACS (4 domains)~\cite{DBLP:conf/iccv/LiYSH17_pacs}, OfficeHome (4 domains)~\cite{DBLP:conf/cvpr/VenkateswaraECP17_officehome} and Office31 (3 domains)~\cite{DBLP:conf/eccv/SaenkoKFD10_office31}, wherein each client's data is exclusively sampled from a single domain.
    \item Pathological Label Skew: We group five visual classification datasets (Food101~\cite{DBLP:conf/eccv/BossardGG14_food101}, Caltech101~\cite{DBLP:conf/cvpr/LiFP04_caltech101}, Flowers102~\cite{DBLP:conf/icvgip/NilsbackZ08_flower102}, OxfordPets~\cite{DBLP:conf/cvpr/ParkhiVZJ12_oxfordpets} and DTD~\cite{DBLP:conf/cvpr/CimpoiMKMV14_dtd}) as a single-domain CLIP benchmark, randomly assigning distinct, non-overlapping classes to each client. 
    \item Dirichlet Imbalance: We partition CIFAR-10~\cite{krizhevsky2009learning} and CIFAR-100~\cite{krizhevsky2009learning} among clients using a Dirichlet distribution with Dirichlet coefficients $\beta \in \{0.1, 0.3, 0.5\}$ to simulate varying degrees of label-distribution heterogeneity.
\end{itemize}

\subsubsection{Baselines.}
We compare our FedIA with the following methods:
\begin{itemize}
    \item Zero-Shot CLIP (ZS-CLIP) \cite{DBLP:conf/icml/RadfordKHRGASAM21_CLIP}: CLIP relies on manually crafted text prompt templates to evaluate and generate the model's initial performance without the need for downstream fine-tuning.
    \item PromptFL \cite{DBLP:journals/tmc/GuoGWTX24_promptfl}: A prompt-based federated learning method that uses a federated averaging mechanism to learn a unified text prompt across clients.
    \item FedMaPLe \cite{DBLP:conf/cvpr/KhattakR0KK23_MaPLe}: It directly extends the MaPLe framework into the FL setting to enable distributed multi-modal prompt tuning.
    \item FedOTP \cite{DBLP:conf/cvpr/LiHW024_FedOTP}: It enhances personalization by employing Unbalanced Optimal Transport to accurately align both global and local prompts with the most relevant local visual features.
    \item FedPGP \cite{DBLP:conf/icml/CuiLW024_FedPGP}: It utilizes low-rank adaptation and a prompt-wise contrastive loss to balance generalization and client-specific personalization in heterogeneous data.
    \item FedTPG \cite{DBLP:conf/iclr/QiuLMGLPL24_FedTPG}: A unified prompt generation network conditioned on task-related textual information, which generates context-aware prompts for adapting vision-language models across decentralized clients.
    \item FedPHA \cite{DBLP:conf/icml/Fang0WYY25_FedPHA}: It combines fixed-length global prompts with variable-length local prompts, utilizing an SVD-based projection mechanism to resolve conflicts between global and local knowledge.
    \item pFedMMA \cite{DBLP:conf/iclr/GhiasvandAP26_pfedmma}: A personalized federated fine-tuning method that inserts lightweight multimodal adapters into the upper layers of a frozen vision-language model. Each adapter has modality-specific down/up projections and a shared cross-modal projection.
\end{itemize}

\subsubsection{Implementation details.}
We use CLIP with a ViT-B/16 \cite{DBLP:conf/iclr/DosovitskiyB0WZ21_ViT} backbone in all experiments.
The prompt length is 16, and prompts are randomly initialized from a normal distribution.
We set the local epoch to $E=1$ and the communication round to $R=50$, except for CIFAR-10 and CIFAR-100 where $R=25$.
The learning rate is 0.001, and the training batch size is 32.
The number of clients varies across the three federated settings. 
For PACS, OfficeHome, and Office31, we assign two clients to each domain by partitioning its data into two subsets, resulting in 8, 8, and 6 clients, respectively. This configuration preserves domain heterogeneity across clients while allowing multiple clients to share the same domain. 
For the CLIP benchmark datasets, including Food101, Caltech101, Flowers102, OxfordPets, and DTD, we use $N=10$ clients with disjoint class subsets to simulate pathological label skew. 
For CIFAR-10 and CIFAR-100, we set $N=100$, with client data partitioned according to a Dirichlet distribution to model heterogeneous label distributions. 
All experiments are conducted on an NVIDIA L20 GPU, and the reported results are averaged over three runs with different random seeds. 

%%%%%%%%%%%%%%%%%%%%%%%%%%%%%%%%% pacs
\begin{table*}[pos=t]
\centering
\caption{Comparison of different methods on PACS under $k=49$. Results are reported as mean $\pm$ standard deviation of accuracy (\%) over three random seeds. Avg denotes the average accuracy across all domains, and the best average results are shown in bold.}
\label{tab:domain_pacs}
\begin{tabular}{l ccccc}
\toprule
\multirow{2}{*}{\textbf{Methods}}
& \multicolumn{5}{c}{\textbf{PACS}}  \\
\cmidrule(lr){2-6}
& A & C & P & S & \textbf{Avg}   \\
\midrule
ZS-CLIP
& $97.26\pm0.00$ & $99.11\pm0.00$ & $99.95\pm0.00$ & $87.77\pm0.00$ & $96.02\pm0.00$
\\

PromptFL
& $98.63\pm0.14$ & $99.24\pm0.05$ & $99.92\pm0.03$ & $94.47\pm0.40$ & $98.06\pm0.11$
\\

FedMaPLe
& $98.47\pm0.21$ & $99.11\pm0.19$ & $99.90\pm0.04$ & $95.43\pm0.06$ & $98.23\pm0.12$
\\

FedOTP
& $98.43\pm0.08$ & $99.18\pm0.09$ & $99.73\pm0.07$ & $96.42\pm0.13$ & $98.44\pm0.03$
\\

FedPGP
& $98.97\pm0.08$ & $99.59\pm0.08$ & $99.95\pm0.01$ & $96.65\pm0.10$ & $98.79\pm0.03$
\\

FedTPG
& $98.47\pm0.07$ & $99.17\pm0.09$ & $99.89\pm0.07$ & $94.45\pm0.11$ & $98.00\pm0.03$
\\

FedPHA
& $99.19\pm0.02$ & $99.65\pm0.03$ & $99.89\pm0.02$ & $97.14\pm0.11$ & $98.97\pm0.02$
\\

pFedMMA
& $97.70\pm0.75$ & $99.33\pm0.07$ & $99.88\pm0.08$ & $99.07\pm0.13$ & $99.00\pm0.17$
\\
\midrule
\textbf{FedIA}
& $99.25\pm0.02$ & $99.74\pm0.05$ & $99.85\pm0.04$ & $97.73\pm0.08$ & $\mathbf{99.14\pm0.02}$
\\
\bottomrule
\end{tabular}
\end{table*}
%%%%%%%%%%%%%%%%%%%%%%%%%%%%%%%%%%%%%%%%%%

%%%%%%%%%%%%%%%%%%%%%%%%%%%%%%%%%% officehome
\begin{table*}[pos=t]
\centering
\caption{Comparison of different methods on OfficeHome under $k=49$. Results are reported as mean $\pm$ standard deviation of accuracy (\%) over three random seeds. Avg denotes the average accuracy across all domains, and the best average results are shown in bold.}
\label{tab:domain_officehome}
\begin{tabular}{l ccccc}
\toprule
\multirow{2}{*}{\textbf{Methods}}
& \multicolumn{5}{c}{\textbf{OfficeHome}} \\
\cmidrule(lr){2-6}
& A & C & P & R & \textbf{Avg} \\
\midrule
ZS-CLIP
& $84.45\pm0.00$ & $66.14\pm0.00$ & $89.50\pm0.00$ & $89.55\pm0.00$ & $82.41\pm0.00$ \\

PromptFL
& $87.13\pm0.75$ & $75.28\pm0.33$ & $94.74\pm0.44$ & $92.48\pm0.26$ & $87.41\pm0.18$ \\

FedMaPLe
& $85.67\pm0.71$ & $77.38\pm0.18$ & $94.58\pm0.30$ & $92.62\pm0.28$ & $87.56\pm0.29$ \\

FedOTP
& $80.50\pm1.08$ & $76.92\pm0.21$ & $92.09\pm0.17$ & $88.02\pm0.08$ & $84.38\pm0.17$ \\

FedPGP
& $88.89\pm0.65$ & $78.32\pm0.64$ & $95.48\pm0.37$ & $93.92\pm0.37$ & $89.15\pm0.22$ \\

FedTPG
& $85.69\pm0.48$ & $73.47\pm0.33$ & $93.95\pm0.15$ & $92.66\pm0.27$ & $86.44\pm0.17$ \\

FedPHA
& $87.78\pm0.34$ & $79.61\pm0.35$ & $95.83\pm0.19$ & $93.66\pm0.14$ & $89.22\pm0.11$ \\

pFedMMA
& $89.36\pm0.44$ & $77.77\pm0.05$ & $93.58\pm0.56$ & $92.46\pm0.31$ & $88.29\pm0.28$ \\

\midrule
\textbf{FedIA}
& $88.67\pm0.98$ & $81.46\pm0.14$ & $95.74\pm0.33$ & $93.18\pm0.15$ & $\mathbf{89.76\pm0.27}$ \\
\bottomrule
\end{tabular}
\end{table*}
%%%%%%%%%%%%%%%%%%%%%%%%%%%%%%%%%%

%%%%%%%%%%%%%%%%%%%%%%%%% office31
\begin{table*}[pos=t]
\centering
\caption{Comparison of different methods on Office31 under $k=49$. Results are reported as mean $\pm$ standard deviation of accuracy (\%) over three random seeds. Avg denotes the average accuracy across all domains, and the best average results are shown in bold.}
\label{tab:domain_office31}
\begin{tabular}{l cccc}
\toprule
\multirow{2}{*}{\textbf{Methods}}
& \multicolumn{4}{c}{\textbf{Office31}}
\\
\cmidrule(lr){2-5}
& A & D & W & \textbf{Avg}
\\
\midrule
ZS-CLIP
& $81.15\pm0.00$ & $71.63\pm0.00$ & $73.15\pm0.00$ & $75.31\pm0.00$
\\

PromptFL
& $88.43\pm0.08$ & $86.70\pm2.01$ & $91.18\pm1.05$ & $88.77\pm0.73$
\\

FedMaPLe
& $88.60\pm0.24$ & $88.59\pm1.61$ & $91.06\pm0.71$ & $89.41\pm0.43$
\\

FedOTP
& $86.33\pm0.47$ & $89.36\pm1.87$ & $93.87\pm0.71$ & $89.85\pm0.52$
\\

FedPGP
& $90.17\pm0.43$ & $91.09\pm1.97$ & $95.40\pm0.33$ & $92.22\pm0.67$
\\

FedTPG
& $86.65\pm0.33$ & $84.19\pm0.97$ & $87.82\pm1.75$ & $86.22\pm0.65$
\\

FedPHA
& $90.37\pm0.46$ & $95.73\pm0.84$ & $97.56\pm0.88$ & $94.55\pm0.02$
\\

pFedMMA
& $89.78\pm0.54$ & $94.65\pm2.02$ & $95.19\pm0.81$ & $93.21\pm0.67$
\\
\midrule
\textbf{FedIA}
& $90.92\pm0.28$ & $96.10\pm0.84$ & $98.20\pm0.66$ & $\mathbf{95.08\pm0.48}$
\\
\bottomrule
\end{tabular}
\end{table*}
%%%%%%%%%%%%%%%%%%%%%%%%%%%%%

\subsection{Performance Analysis}
We first evaluate FedIA under three heterogeneous federated settings: domain shift, pathological label skew, and Dirichlet imbalance, covering variations in both visual domains and client label distributions. 
We set $k=49$ as the default number of selected patches in FedIA. For the ViT-B/16 backbone, this retains 25\% of the 196 visual patches. 
This selection ratio allows the model to focus on patches with high category relevance while retaining diverse visual cues for instance-aware text representation generation. 

\textbf{Results under domain shift.}
To evaluate the effectiveness of FedIA under domain shift, we conduct experiments on PACS, OfficeHome, and Office31, partitioning the data within each domain between two clients using a Dirichlet distribution with coefficient $\beta = 0.5$ to simulate client heterogeneity. We analyze the average performance of clients within each domain and the overall average performance of all clients.
Table~\ref{tab:domain_pacs} reports an average accuracy of 99.14\% for FedIA on PACS, which is higher than those of all compared methods. 
For OfficeHome, the results in Table~\ref{tab:domain_officehome} show that FedIA reaches an average accuracy of 89.76\%. It achieves a particularly notable result on the Clipart domain, reaching 81.46\% and exceeding FedPHA by 1.85 percentage points.
On Office31, FedIA obtains the highest average accuracy of 95.08\%, as reported in Table~\ref{tab:domain_office31}. It also achieves the best performance on all three domains.
Overall, FedIA achieves the highest average accuracy on PACS, OfficeHome, and Office31, reaching 99.14\%, 89.76\%, and 95.08\%, respectively. 
These results are consistent with our design motivation: instance-aware text representations enable adaptation to sample-specific visual cues, while adaptive global--local fusion balances personalized information with semantic knowledge shared across clients.

%%%%%%%%%%%%%%%%%%%%%%%%%%%%%%%%%%%%%%%% domain acc
\begin{figure*}[pos=t]
    \centering
    \begin{subfigure}[t]{\dimexpr\linewidth/3\relax}
        \centering
        \includegraphics[width=\linewidth]{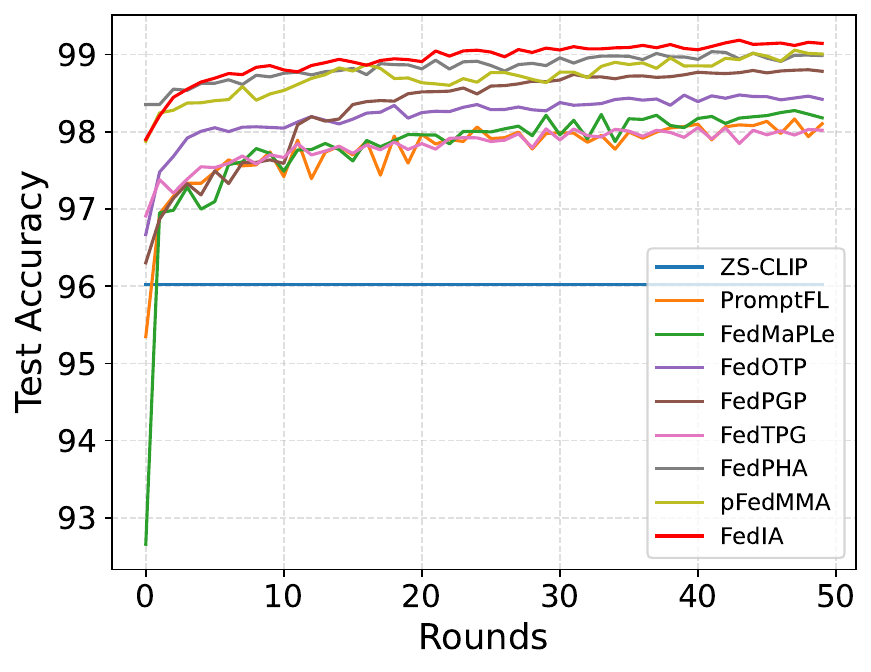}
        \caption{PACS}
        \label{fig:acc_pacs}
    \end{subfigure}\hfill%
    \begin{subfigure}[t]{\dimexpr\linewidth/3\relax}
        \centering
        \includegraphics[width=\linewidth]{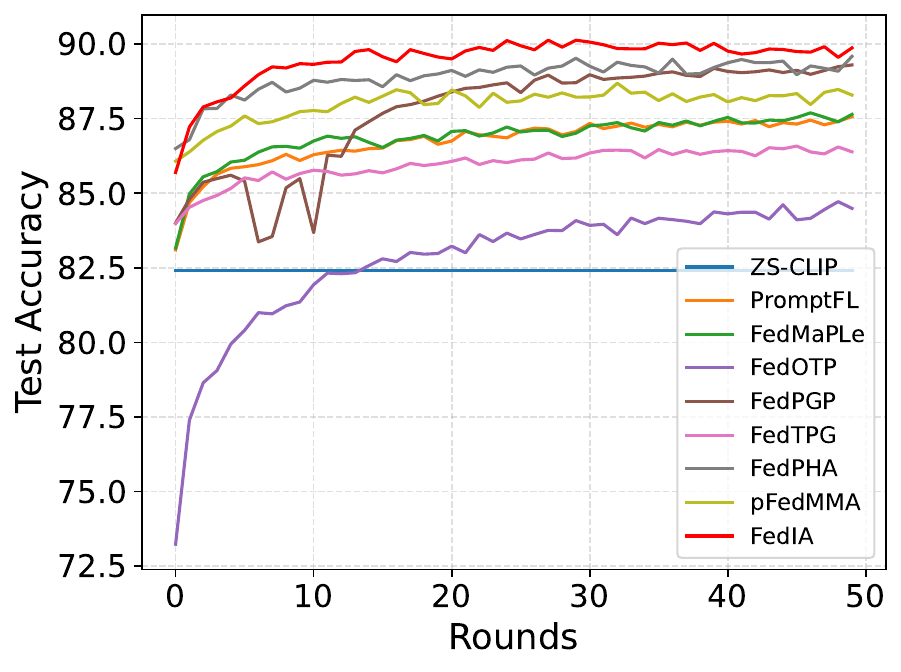}
        \caption{OfficeHome}
        \label{fig:acc_officehome}
    \end{subfigure}\hfill%
    \begin{subfigure}[t]{\dimexpr\linewidth/3\relax}
        \centering
        \includegraphics[width=\linewidth]{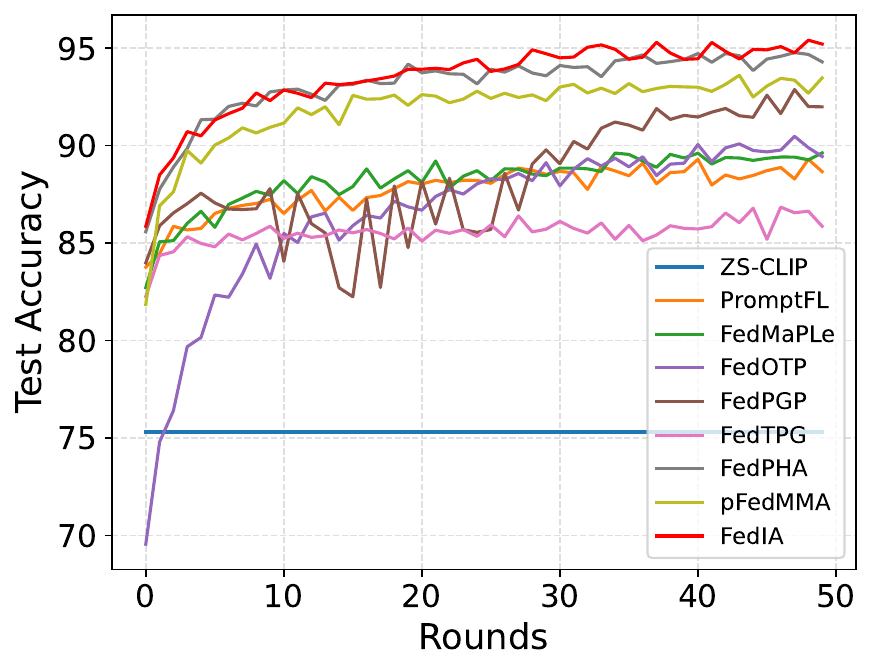}
        \caption{Office31}
        \label{fig:acc_office31}
    \end{subfigure}

    \caption{Test accuracy over communication rounds on PACS, OfficeHome, and Office31 with $k=49$.}
    \label{fig:convergence_domain_acc}
\end{figure*}
%%%%%%%%%%%%%%%%%%%%%%%%%%%%%%%%%%%%%%%%%%%%

%%%%%%%%%%%%%%%%%%%%%%%%%% clip five 
\begin{table*}[pos=t]
    \centering
    \caption{Comparison of different algorithms across pathological label skew datasets under $k=49$. Results are reported as mean $\pm$ standard deviation of accuracy (\%) over three random seeds.}
    \label{tab:clip5}
        \begin{tabular}{l ccccc}
            \toprule
            \textbf{Methods}
            & \textbf{Food101}
            & \textbf{Caltech101}
            & \textbf{Flowers102}
            & \textbf{OxfordPets}
            & \textbf{DTD} \\
            \midrule
            ZS-CLIP
            & $85.26\pm0.00$ & $91.93\pm0.00$ & $67.88\pm0.00$
            & $89.05\pm0.00$ & $43.33\pm0.00$
            \\

            PromptFL
            & $86.76\pm0.19$ & $93.42\pm0.15$ & $87.55\pm2.07$
            & $93.43\pm0.27$ & $56.08\pm0.57$
            \\

            FedMaPLe
            & $86.66\pm0.22$ & $93.81\pm0.15$ & $87.60\pm1.61$
            & $93.48\pm0.61$ & $55.44\pm0.78$
            \\

            FedOTP
            & $92.97\pm0.18$ & $98.13\pm0.18$ & $98.55\pm0.14$
            & $98.70\pm0.09$ & $90.53\pm0.40$
            \\

            FedPGP
            & $87.88\pm0.78$ & $94.96\pm1.10$ & $96.97\pm1.00$
            & $94.68\pm0.57$ & $62.24\pm2.34$
            \\

            FedTPG
            & $86.53\pm0.45$ & $93.86\pm0.46$ & $80.04\pm4.42$
            & $93.07\pm0.11$ & $55.17\pm0.75$
            \\

            FedPHA
            & $96.37\pm0.05$ & $98.84\pm0.05$ & $\mathbf{99.31\pm0.06}$
            & $99.07\pm0.19$ & $\mathbf{91.94\pm0.61}$
            \\
            pFedMMA
            & $96.03\pm0.18$ & $98.57\pm0.13$ & $95.74\pm1.50$
            & $98.69\pm0.15$ & $87.32\pm1.28$
            \\
            \midrule
            \textbf{FedIA}
            & $\mathbf{96.58\pm0.04}$ & $\mathbf{99.13\pm0.10}$ & $99.25\pm0.03$
            & $\mathbf{99.28\pm0.08}$ & $91.69\pm0.27$
            \\
            \bottomrule
        \end{tabular}
\end{table*}
%%%%%%%%%%%%%%%%%%%%%%%%%%

\textbf{Results under pathological label skew.}
Table~\ref{tab:clip5} reports the results on five single-domain CLIP benchmarks, including Food101, Caltech101, Flowers102, OxfordPets, and DTD, under the 16-shot setting. 
Each dataset is partitioned among 10 clients with non-overlapping class subsets, resulting in substantial label heterogeneity and limited class coverage at each client. 
FedIA achieves the best performance on Food101, Caltech101, and OxfordPets, reaching 96.58\%, 99.13\%, and 99.28\%, respectively.  
The small standard deviations across three random seeds also suggest stable performance under this setting.
Overall, these results indicate that client-level or globally shared prompts alone have limited ability to address severe class heterogeneity. By constructing instance-aware representations, FedIA captures both client-specific semantics and visual variations among individual samples, allowing personalization to account for the characteristics of each input. Meanwhile, its adaptive global--local fusion retains shared semantic knowledge while incorporating instance-specific local information, thereby improving personalization without sacrificing cross-client generalization.

%%%%%%%%%%%%%%%%%%%%%%%%%%% cifar
\begin{table*}[t]
    \centering
    \caption{Performance comparison on CIFAR-10 and CIFAR-100 under Dirichlet partitioning with $\beta \in \{0.1,0.3,0.5\}$ and $k=49$. Results are reported as mean $\pm$ standard deviation of accuracy (\%) over three random seeds.}
    \label{tab:cifar}
        \begin{tabular}{l ccc ccc}
            \toprule
            \multirow{2}{*}{\textbf{Methods}}
            & \multicolumn{3}{c}{\textbf{CIFAR-10}}
            & \multicolumn{3}{c}{\textbf{CIFAR-100}} \\
            \cmidrule(lr){2-4}
            \cmidrule(lr){5-7}
            & $\beta=0.1$ & $\beta=0.3$ & $\beta=0.5$ 
            & $\beta=0.1$ & $\beta=0.3$ & $\beta=0.5$
            \\ 
            \midrule
            ZS-CLIP
            & $87.84\pm0.38$ & $88.05\pm0.11$ & $87.84\pm0.15$
            & $64.94\pm0.13$ & $64.89\pm0.02$ & $64.90\pm0.04$ \\
        
            PromptFL
            & $89.68\pm0.67$ & $91.70\pm0.18$ & $91.80\pm0.38$
            & $70.96\pm0.15$ & $72.45\pm0.14$ & $72.89\pm0.15$ \\
        
            FedMaPLe
            & $92.57\pm0.44$ & $93.44\pm0.60$ & $93.58\pm0.97$
            & $74.61\pm0.43$ & $75.63\pm0.26$ & $75.92\pm0.36$ \\
        
            FedOTP
            & $96.95\pm0.31$ & $95.07\pm0.30$ & $93.55\pm0.38$
            & $86.47\pm0.18$ & $79.05\pm0.28$ & $75.21\pm0.33$ \\
        
            FedPGP
            & $94.67\pm0.14$ & $94.20\pm0.37$ & $93.35\pm0.73$
            & $86.44\pm0.05$ & $79.09\pm0.51$ & $75.81\pm0.43$ \\
        
            FedTPG
            & $89.20\pm0.25$ & $91.14\pm0.44$ & $91.29\pm0.58$
            & $69.40\pm0.39$ & $70.10\pm0.43$ & $70.08\pm0.54$ \\
        
            FedPHA
            & $97.35\pm0.29$ & $95.58\pm0.27$ & $94.15\pm0.33$
            & $87.35\pm0.10$ & $79.86\pm0.37$ & $76.10\pm0.41$ \\
        
            pFedMMA
            & $97.16\pm0.27$ & $95.43\pm0.10$ & $93.82\pm0.26$
            & $87.54\pm0.22$ & $79.75\pm0.54$ & $76.13\pm0.23$ \\
        
            \midrule
        
            \textbf{FedIA}
            & $\mathbf{97.40\pm0.19}$ & $\mathbf{95.73\pm0.19}$ & $\mathbf{94.21\pm0.27}$
            & $\mathbf{88.28\pm0.34}$ & $\mathbf{81.10\pm0.46}$ & $\mathbf{77.57\pm0.61}$ \\
            \bottomrule
        \end{tabular}
\end{table*}
%%%%%%%%%%%%%%%%%%%%%%%%%%%%%%%

%%%%%%%%%%%%%%%%%%%%%%%%%%%%%%%%%%%%%%%% cifar10 acc
\begin{figure*}[pos=t]
    \centering
    \begin{subfigure}[t]{\dimexpr\linewidth/3\relax}
        \centering
        \includegraphics[width=\linewidth]{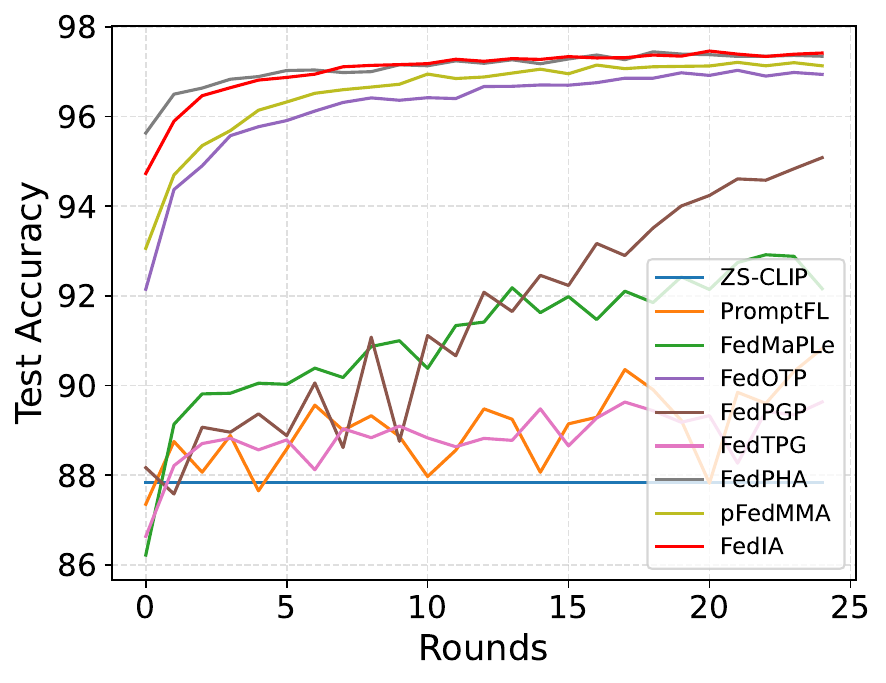}
        \caption{$\beta=0.1$}
        \label{fig:acc_cifar10_01}
    \end{subfigure}\hfill%
    \begin{subfigure}[t]{\dimexpr\linewidth/3\relax}
        \centering
        \includegraphics[width=\linewidth]{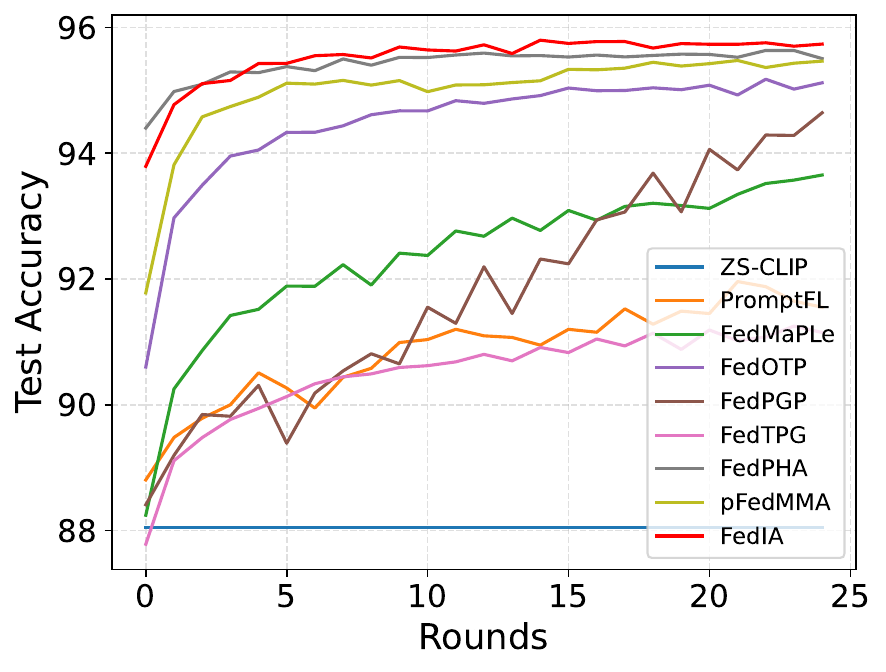}
        \caption{$\beta=0.3$}
        \label{fig:acc_cifar10_03}
    \end{subfigure}\hfill%
    \begin{subfigure}[t]{\dimexpr\linewidth/3\relax}
        \centering
        \includegraphics[width=\linewidth]{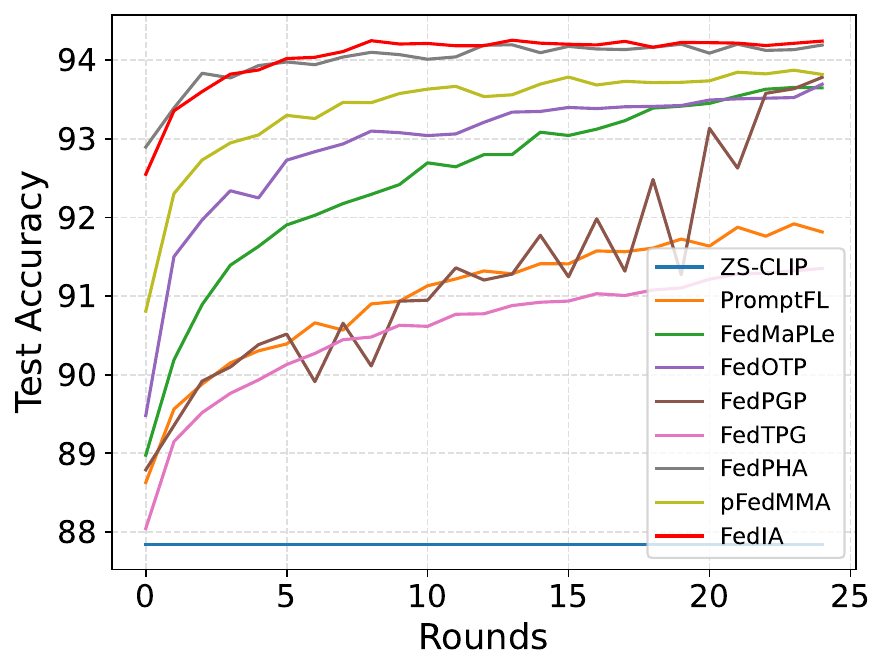}
        \caption{$\beta=0.5$}
        \label{fig:acc_cifar10_05}
    \end{subfigure}

    \caption{Test accuracy over communication rounds on CIFAR-10 under Dirichlet partitioning with $\beta \in \{0.1,0.3,0.5\}$ and $k=49$.}
    \label{fig:convergence_cifar10_acc}
\end{figure*}
%%%%%%%%%%%%%%%%%%%%%%%%%%%%%%%%%%%%%%%%%%%%

%%%%%%%%%%%%%%%%%%%%%%%%%%%%%%%%%%%%%%%% cifar100 acc
\begin{figure*}[pos=t]
    \centering
    \begin{subfigure}[t]{\dimexpr\linewidth/3\relax}
        \centering
        \includegraphics[width=\linewidth]{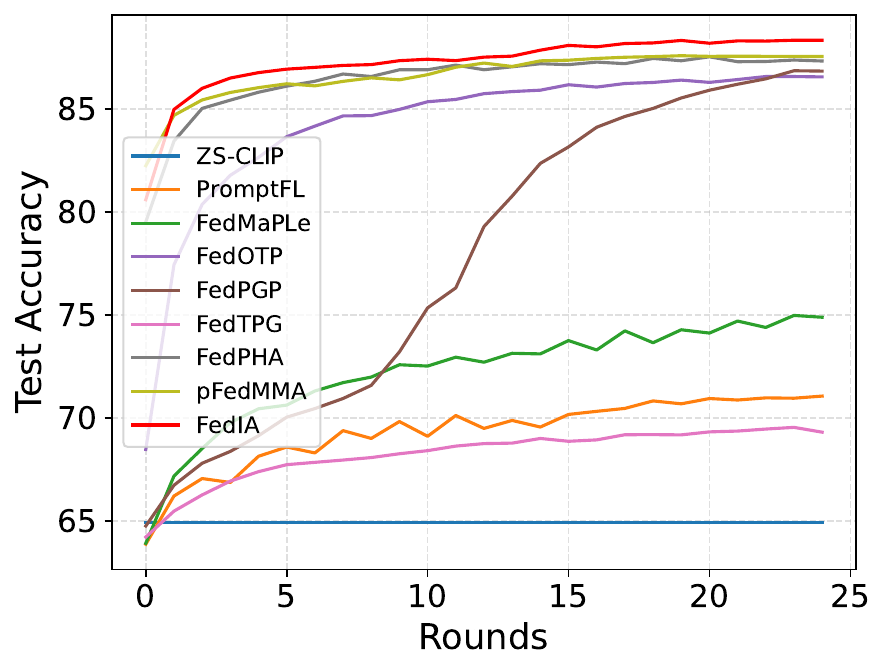}
        \caption{$\beta=0.1$}
        \label{fig:acc_cifar100_01}
    \end{subfigure}\hfill%
    \begin{subfigure}[t]{\dimexpr\linewidth/3\relax}
        \centering
        \includegraphics[width=\linewidth]{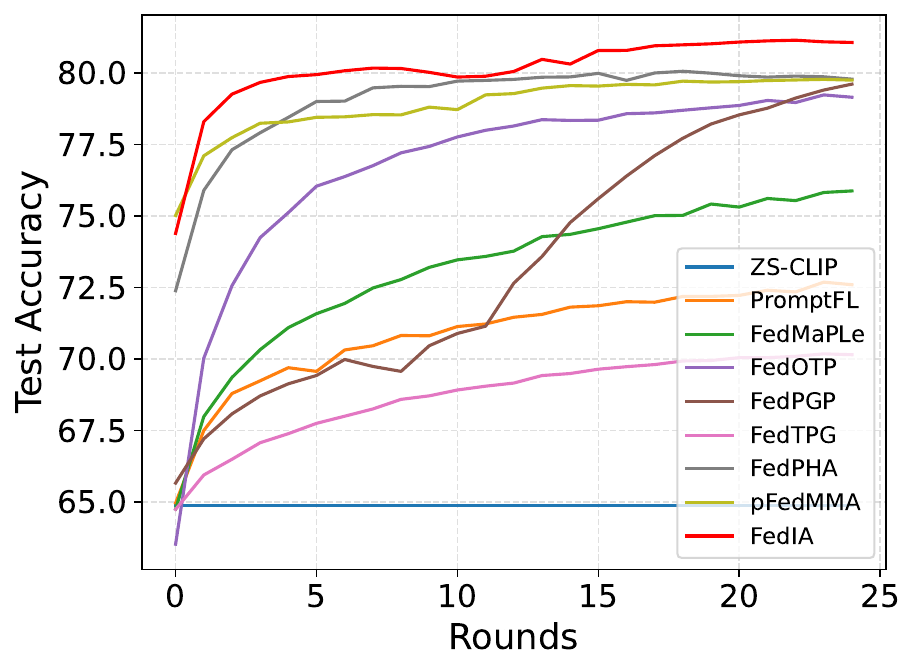}
        \caption{$\beta=0.3$}
        \label{fig:acc_cifar100_03}
    \end{subfigure}\hfill%
    \begin{subfigure}[t]{\dimexpr\linewidth/3\relax}
        \centering
        \includegraphics[width=\linewidth]{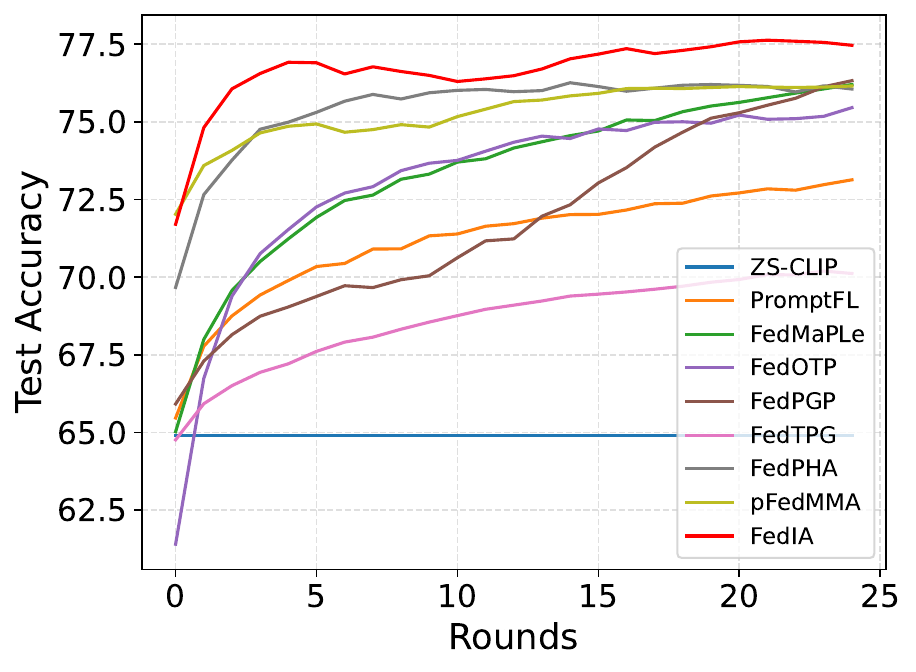}
        \caption{$\beta=0.5$}
        \label{fig:acc_cifar100_05}
    \end{subfigure}

    \caption{Test accuracy over communication rounds on CIFAR-100 under Dirichlet partitioning with $\beta \in \{0.1,0.3,0.5\}$ and $k=49$.}
    \label{fig:convergence_cifar100_acc}
\end{figure*}
%%%%%%%%%%%%%%%%%%%%%%%%%%%%%%%%%%%%%%%%%%%%

\textbf{Results under Dirichlet imbalance.}
We evaluate FedIA on CIFAR-10 and CIFAR-100 using Dirichlet partitions with $\beta \in \{0.1,0.3,0.5\}$ to examine its performance under varying degrees of label heterogeneity. Smaller values of $\beta$ generally produce more skewed client-level class distributions.
Table~\ref{tab:cifar} shows that FedIA achieves the highest average accuracy among the compared methods on both datasets for all three values of $\beta$.
On CIFAR-10, FedIA obtains accuracies of 97.40\%, 95.73\%, and 94.21\% for $\beta=0.1$, $0.3$, and $0.5$, respectively. The corresponding accuracies on CIFAR-100 are 88.28\%, 81.10\%, and 77.57\%.
The improvements are more pronounced on CIFAR-100, where FedIA exceeds the strongest baseline in each setting by 0.74, 1.24, and 1.44 percentage points, respectively. In particular, at $\beta=0.5$, FedIA improves the accuracy from 76.13\% achieved by pFedMMA to 77.57\%. With more categories and visually similar classes, CIFAR-100 places greater demands on the model's ability to capture discriminative visual details. These results support the effectiveness of combining instance-aware local text representations with adaptive global--local fusion under different levels of label heterogeneity.

Overall, the results across the three heterogeneous federated settings demonstrate the effectiveness of FedIA in generating instance-aware local text representations and adaptively fusing them with shared global representations.

\subsection{Ablation Study}
We conduct ablation studies to assess the individual contributions and combined effectiveness of Instance-aware Local Representation Generation (ILRG) and Adaptive Global--Local Representation Fusion (AGLRF).

Table~\ref{tab:ablation_office31} presents the results on Office31. Both ILRG and AGLRF independently improve the average accuracy over the baseline, demonstrating the benefit of each component. Their combination further increases performance, achieving the highest average accuracies of 94.99\% and 95.08\% under $k=25$ and $k=49$, respectively. The consistent improvements across both patch selection settings support the complementary roles of the ILRG and AGLRF.
Table~\ref{tab:ablation_cifar_officehome} further evaluates their contributions on OfficeHome, CIFAR-10 and CIFAR-100. Using ILRG alone consistently improves the baseline under both $k=25$ and $k=49$. Specifically, with $k=49$, ILRG increases accuracy by 2.15, 2.19, and 4.14 percentage points on the three datasets, respectively. These gains support the effectiveness of generating instance-aware local text representations from selected visual patches, allowing local representations to incorporate fine-grained visual information specific to each input.
Using AGLRF alone also yields consistent gains, with an improvement of 3.28 percentage points over the baseline on CIFAR-100 under $k=49$. This result supports adaptively adjusting the contributions of personalized local information and globally shared semantics. When both modules are enabled, FedIA achieves the highest accuracy among the evaluated configurations on all three datasets under both values of $k$, further improving upon either module alone.

Together, these results demonstrate the complementary benefits of ILRG and AGLRF: ILRG enriches local text representations with instance-specific visual evidence, while AGLRF adaptively integrates these representations with shared semantic knowledge.

%%%%%%%%%%%%%%%%%%%%%%%%%%%%%%%%%%%%%%%%%%%%%
\begin{table*}[t]
    \centering
    \caption{Ablation study of FedIA on Office31 under $k=25$ and $k=49$. $\Delta$ denotes the accuracy improvement over the baseline in percentage points.}
    \label{tab:ablation_office31}
        \begin{tabular}{ccc cc cc cc cc}
            \toprule
            $k$
            & \textbf{ILRG}
            & \textbf{AGLRF}
            & \textbf{A} & $\Delta$
            & \textbf{D} & $\Delta$
            & \textbf{W} & $\Delta$
            & \textbf{Avg} & $\Delta$ \\
            \midrule
            \multirow{4}{*}{25}
            & & 
            & $88.43$ & --
            & $86.70$ & --
            & $91.18$ & --
            & $88.77$ & -- \\

            & $\checkmark$ &
            & $90.70$ & $\uparrow 2.27$
            & $95.82$ & $\uparrow 9.12$
            & $97.67$ & $\uparrow 6.49$
            & $94.73$ & $\uparrow 5.96$ \\

            & & $\checkmark$
            & $90.40$ & $\uparrow 1.97$
            & $95.36$ & $\uparrow 8.66$
            & $97.65$ & $\uparrow 6.47$
            & $94.47$ & $\uparrow 5.70$ \\

            & $\checkmark$ & $\checkmark$
            & $\mathbf{90.94}$ & $\uparrow \mathbf{2.51}$
            & $\mathbf{95.95}$ & $\uparrow \mathbf{9.25}$
            & $\mathbf{98.08}$ & $\uparrow \mathbf{6.90}$
            & $\mathbf{94.99}$ & $\uparrow \mathbf{6.22}$ \\
            \midrule
            \multirow{4}{*}{49}
            & &
            & $88.43$ & --
            & $86.70$ & --
            & $91.18$ & --
            & $88.77$ & -- \\

            & $\checkmark$ &
            & $90.60$ & $\uparrow 2.17$
            & $95.47$ & $\uparrow 8.77$
            & $98.01$ & $\uparrow 6.83$
            & $94.69$ & $\uparrow 5.92$ \\

            & & $\checkmark$
            & $90.40$ & $\uparrow 1.97$
            & $95.36$ & $\uparrow 8.66$
            & $97.65$ & $\uparrow 6.47$
            & $94.47$ & $\uparrow 5.70$ \\

            & $\checkmark$ & $\checkmark$
            & $\mathbf{90.92}$ & $\uparrow \mathbf{2.49}$
            & $\mathbf{96.10}$ & $\uparrow \mathbf{9.40}$
            & $\mathbf{98.20}$ & $\uparrow \mathbf{7.02}$
            & $\mathbf{95.08}$ & $\uparrow \mathbf{6.31}$ \\
            \bottomrule
        \end{tabular}
\end{table*}
%%%%%%%%%%%%%%%%%%%%%%%%%%%%%%%%%%%%%%%%%%%%%%%%%%%%%%%%%%%%%%%

%%%%%%%%%%%%%%%%%%%%%%%%%%%%%%%%%%%%%%%%%%
\begin{table*}[htbp]
    \centering
    \caption{Ablation study of FedIA on OfficeHome, CIFAR-10, and CIFAR-100 under $k=25$ and $k=49$ with $\beta=0.5$. The first value in each pair is the average accuracy (\%), and the second is the improvement over the baseline in percentage points.}
    \label{tab:ablation_cifar_officehome}
        \begin{tabular}{cc cc cc cc cc cc cc}
            \toprule
            \multirow{2}{*}{\textbf{ILRG}}
            & \multirow{2}{*}{\textbf{AGLRF}}
            & \multicolumn{6}{c}{$k=25$}
            & \multicolumn{6}{c}{$k=49$} \\
            \cmidrule(lr){3-8}
            \cmidrule(lr){9-14}
            & &
            \multicolumn{2}{c}{\textbf{OfficeHome}}
            & \multicolumn{2}{c}{\textbf{CIFAR-10}}
            & \multicolumn{2}{c}{\textbf{CIFAR-100}}
            
            & \multicolumn{2}{c}{\textbf{OfficeHome}}
            & \multicolumn{2}{c}{\textbf{CIFAR-10}}
            & \multicolumn{2}{c}{\textbf{CIFAR-100}} \\
            \midrule
            & 
            & $87.41$ & --
            & $91.80$ & --
            & $72.89$ & --

            & $87.41$ & --
            & $91.80$ & --
            & $72.89$ & -- \\

            $\checkmark$ & 
            & $89.45$ & $\uparrow 2.04$
            & $93.93$ & $\uparrow 2.13$
            & $76.83$ & $\uparrow 3.94$
            
            & $89.56$ & $\uparrow 2.15$
            & $93.99$ & $\uparrow 2.19$
            & $77.03$ & $\uparrow 4.14$ \\

            & $\checkmark$ 
            & $89.57$ & $\uparrow 2.16$
            & $94.19$ & $\uparrow 2.39$
            & $76.17$ & $\uparrow 3.28$

            & $89.57$ & $\uparrow 2.16$
            & $94.19$ & $\uparrow 2.39$
            & $76.17$ & $\uparrow 3.28$ \\

            $\checkmark$ & $\checkmark$ 
            & $\mathbf{89.73}$ & $\uparrow \mathbf{2.32}$
            & $\mathbf{94.22}$ & $\uparrow \mathbf{2.42}$
            & $\mathbf{77.53}$ & $\uparrow \mathbf{4.64}$
            
            & $\mathbf{89.76}$ & $\uparrow \mathbf{2.35}$ 
            & $\mathbf{94.21}$ & $\uparrow \mathbf{2.41}$
            & $\mathbf{77.57}$ & $\uparrow \mathbf{4.68}$ \\
            \bottomrule
        \end{tabular}
\end{table*}
%%%%%%%%%%%%%%%%%%%%%%%%%%%%%%%%%%%%%%%%%%%%

\subsection{Convergence Analysis}
We examine the convergence behavior of FedIA by tracking test accuracy over communication rounds on the domain-shift benchmarks and the CIFAR datasets. 
Figure~\ref{fig:convergence_domain_acc} presents the results on PACS, OfficeHome, and Office31. FedIA achieves substantial accuracy improvements during the early rounds and subsequently maintains competitive performance. On PACS, it converges to an accuracy comparable to the strongest baselines, while on OfficeHome and Office31, it maintains a clearer advantage throughout much of the training process.
Figures~\ref{fig:convergence_cifar10_acc} and~\ref{fig:convergence_cifar100_acc} further report the results under Dirichlet partitions with $\beta \in \{0.1,0.3,0.5\}$. On CIFAR-10, FedIA improves rapidly during the initial rounds and reaches a high accuracy early in training across all three settings. Its subsequent improvements are more gradual, with limited fluctuations as training progresses. Several strong baselines exhibit similar convergence trends, but FedIA consistently finishes with the highest accuracy.
On CIFAR-100, FedIA also shows rapid initial improvements and maintains a clear final advantage across the three Dirichlet settings. 
Overall, these curves show that FedIA learns effectively within the evaluated communication budgets across different heterogeneous settings. Together with the ablation results, they support the effectiveness of instance-aware local representation generation and adaptive global--local fusion.

%%%%%%%%%%%%%%%%%%%%%%%%%%%%% k
\begin{figure*}[pos=t]
    \centering
    \begin{subfigure}[t]{\dimexpr\linewidth/3\relax}
        \centering
        \includegraphics[width=\linewidth]{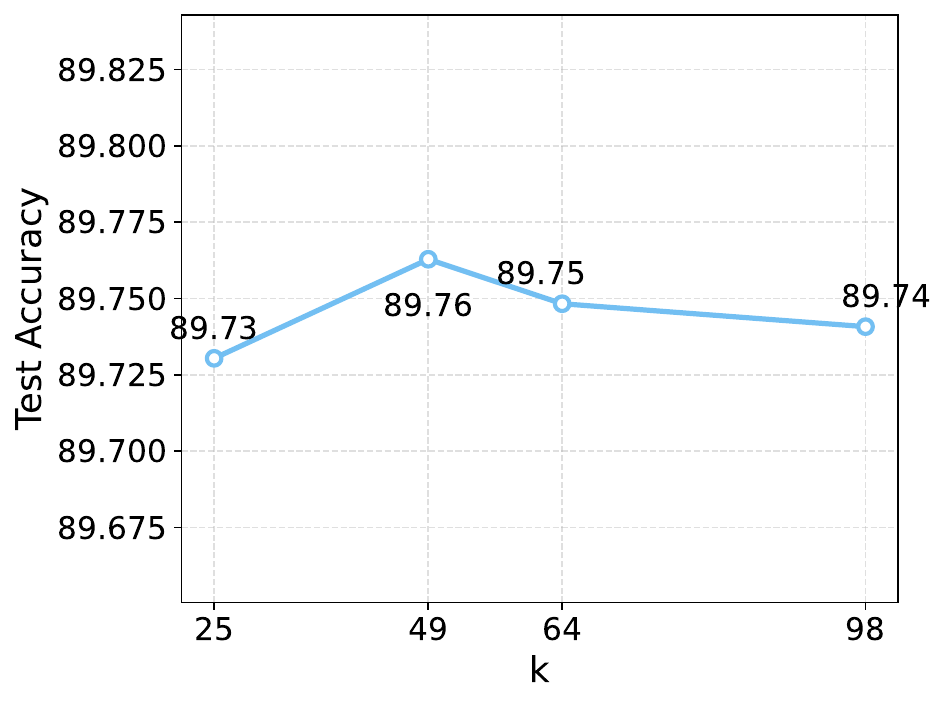}
        \caption{OfficeHome}
        \label{fig:k_officehome}
    \end{subfigure}\hfill%
    \begin{subfigure}[t]{\dimexpr\linewidth/3\relax}
        \centering
        \includegraphics[width=\linewidth]{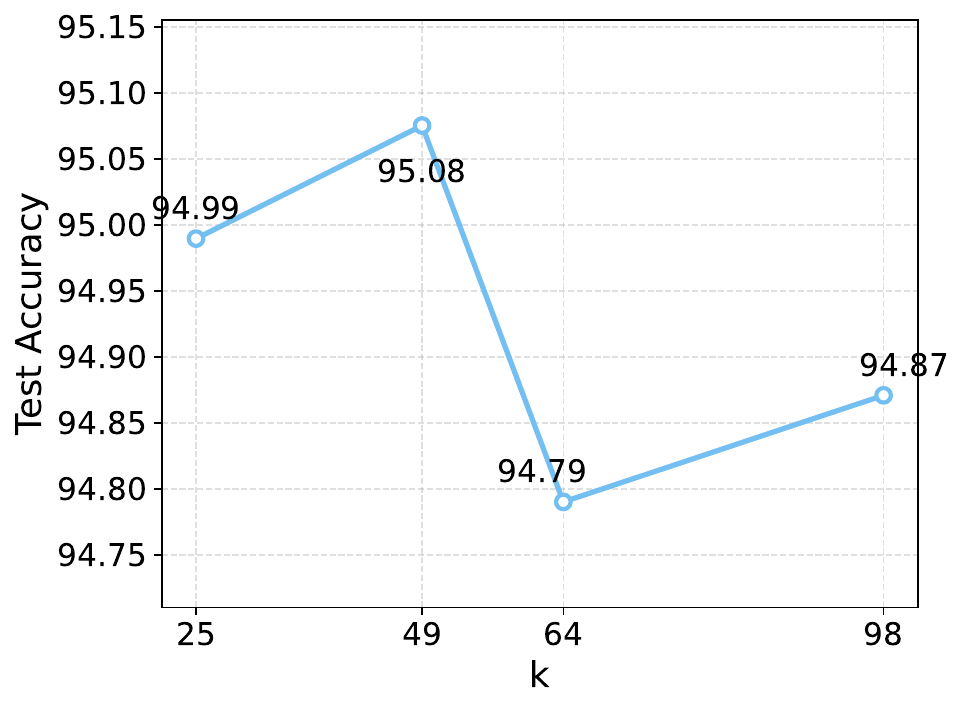}
        \caption{Office31}
        \label{fig:k_office31}
    \end{subfigure}\hfill%
    \begin{subfigure}[t]{\dimexpr\linewidth/3\relax}
        \centering
        \includegraphics[width=\linewidth]{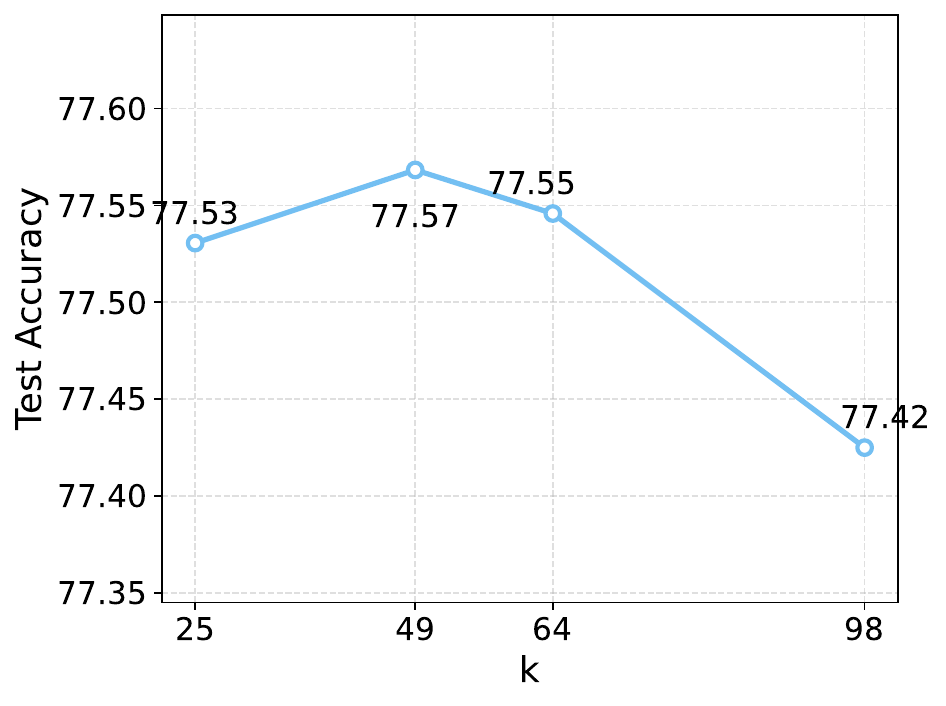}
        \caption{CIFAR-100}
        \label{fig:k_cifar100}
    \end{subfigure}

    \caption{Test accuracy of FedIA on OfficeHome, Office31, and CIFAR-100 under different values of $k$.}
    \label{fig:k}
\end{figure*}
%%%%%%%%%%%%%%%%%%%%%%%%%%%%%%%%%%%%%%%%

%%%%%%%%%%%%%%%%%%%%%%%%%%%%%%%%% shot
\begin{figure*}[pos=t]
    \centering

    \begin{subfigure}[t]{\dimexpr\linewidth/3\relax}
        \centering
        \includegraphics[width=\linewidth]{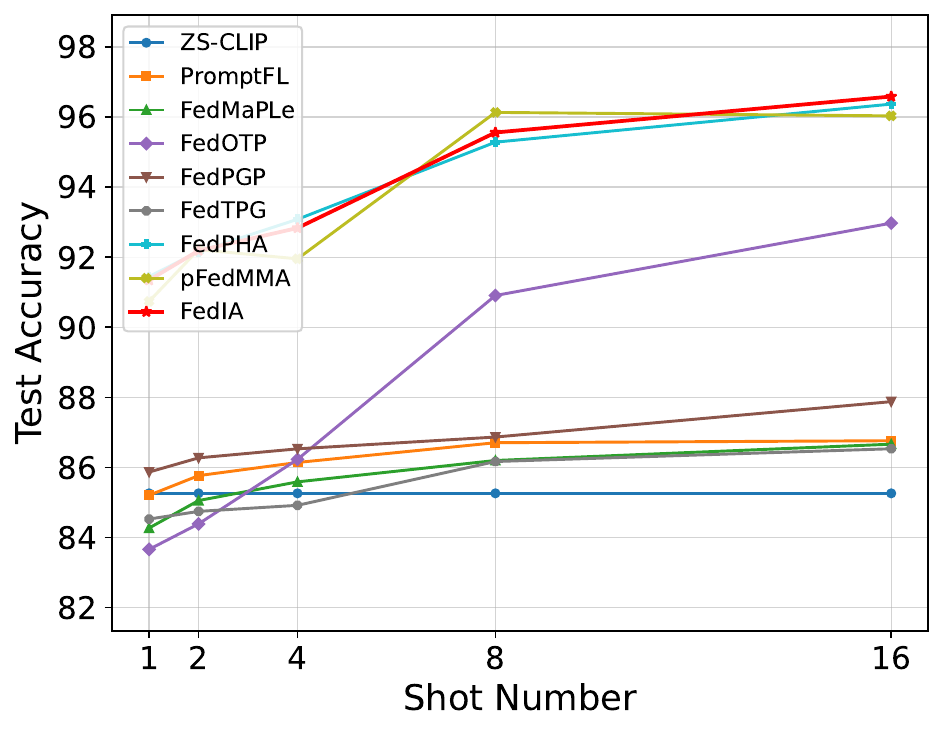}
        \caption{Food101}
    \end{subfigure}\hfill%
    \begin{subfigure}[t]{\dimexpr\linewidth/3\relax}
        \centering
        \includegraphics[width=\linewidth]{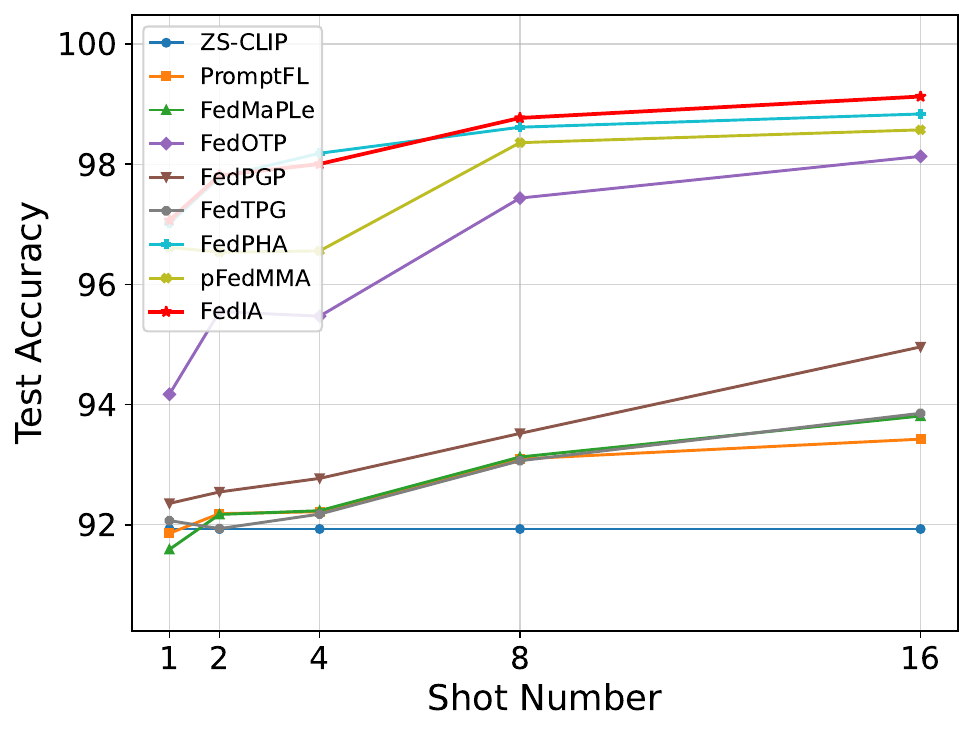}
        \caption{Caltech101}
    \end{subfigure}\hfill%
    \begin{subfigure}[t]{\dimexpr\linewidth/3\relax}
        \centering
        \includegraphics[width=\linewidth]{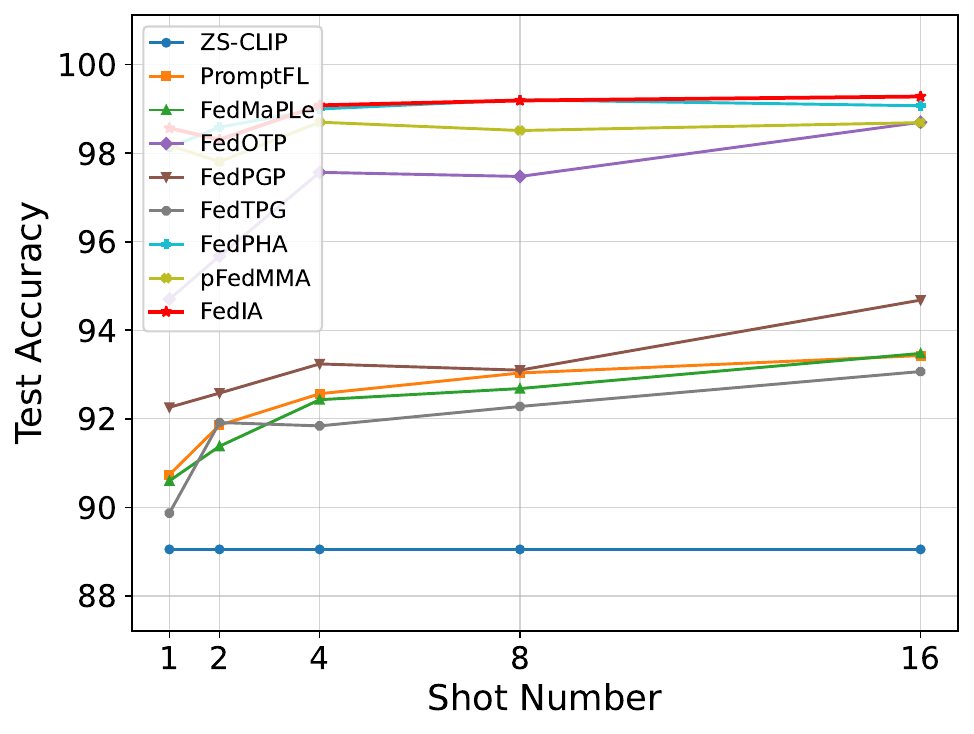}
        \caption{OxfordPets}
    \end{subfigure}
    
    \caption{Test accuracy on Food101, Caltech101, and OxfordPets under different numbers of shots.}
    \label{fig:shot}
\end{figure*}
%%%%%%%%%%%%%%%%%%%%%%%%%%%%%%%%%%%%%%%%%%%%%

\subsection{Sensitivity Analysis}
\textbf{Sensitivity to $k$.}
As shown in Figure~\ref{fig:k}, FedIA maintains relatively stable performance across different values of $k$, indicating limited sensitivity to the number of selected patches. 
On CIFAR-100 with $\beta=0.5$, accuracy varies only slightly from $k=25$ to $k=64$, reaching its highest value of 77.57\% at $k=49$, before decreasing to 77.42\% at $k=98$. 
OfficeHome and Office31 exhibit a similar preference for a moderate selection size, achieving their best accuracies of 89.76\% and 95.08\%, respectively, at $k=49$.
These results suggest that a moderate number of semantically relevant patches provides sufficient visual evidence for instance-aware text representation generation. 
A small $k$ excludes useful category-relevant cues, whereas a large $k$ includes additional patches with lower semantic relevance, reducing the concentration of informative visual content. 
Overall, the small accuracy variations across the evaluated settings support the robustness of FedIA to this hyperparameter and the choice of $k=49$ as the default value.

\textbf{Sensitivity to Shot Number.}
Figure~\ref{fig:shot} reports the performance under different shot numbers on Food101, Caltech101 and OxfordPets.
FedIA achieves gains in low-shot scenarios, suggesting that instance-aware text representation generation can make effective use of limited supervision by incorporating sample-specific visual information. 
As the number of shots increases, its accuracy further improves, showing that the method also benefits from additional labeled examples. 
FedIA remains competitive across the evaluated shot settings and achieves the highest average accuracy at 16 shots on these three datasets.
These results support its effectiveness under varying levels of data availability, from limited-label settings to those with more supervision.

\section{Conclusion}
In this work, we propose FedIA, an FPL framework that learns instance-aware cross-modal representations for adapting vision-language models under heterogeneous federated settings. 
FedIA addresses the limitations of client-level prompts and input-agnostic global--local interactions by selecting class-relevant visual patches, generating image-conditioned local text representations, and adaptively fusing them with globally shared text representations. 
Through this design, FedIA captures fine-grained instance-level visual semantics while preserving transferable global knowledge.  
Extensive experiments under diverse Non-IID settings demonstrate that FedIA generally outperforms representative FPL baselines. Furthermore, ablation studies and sensitivity analyses verify the effectiveness, contribution, and robustness of its key components. Overall, the results show that FedIA provides an effective solution for improving personalized multimodal adaptation in heterogeneous federated environments.

%% Loading bibliography style file
%\bibliographystyle{model1-num-names}
\bibliographystyle{elsarticle-num}
%\bibliographystyle{cas-model2-names}

% Loading bibliography database
% \bibliography{cas-refs}
\bibliography{FedIA_ref}

% Biography
%\bio{}
% Here goes the biography details.
%\endbio

%\bio{pic1}
% Here goes the biography details.
%\endbio

\end{document}